\documentclass[10pt,journal,twocolumn]{IEEEtran}
\usepackage{cite}
\usepackage{url}
\usepackage{ragged2e}
\usepackage{footnote}
\makesavenoteenv{tabular}
\makesavenoteenv{table}
\usepackage{epsfig}
\usepackage{graphicx}
\usepackage{amsmath,amssymb} 
\usepackage{algorithm}
\usepackage{algorithmicx}
\usepackage{algpseudocode}
\usepackage{graphics}
\usepackage{threeparttable}
\usepackage{color}
\usepackage[normalem]{ulem}
\usepackage{multirow}
\usepackage{float}
\usepackage{amsfonts}
\usepackage{bm}
\usepackage{array}
\usepackage{pifont}
\usepackage{diagbox}
\usepackage{rotating}
\usepackage{booktabs}
\usepackage{overpic}
\usepackage{textcomp}
\usepackage{contour}
\usepackage{makecell}

\usepackage{enumitem}
\usepackage{colortbl}
\usepackage[american]{babel}
\usepackage{microtype}
\usepackage{bbding}

\usepackage[pagebackref=false,breaklinks=true,colorlinks, bookmarks=false]{hyperref}

\usepackage{colortbl}
\usepackage{cleveref}

\usepackage{silence}
\graphicspath{{./Imgs/}{./Imgs/authors/}}
\DeclareGraphicsExtensions{.pdf,.jpg,.png}

\def\ie{\textit{i.e.}}
\def\eg{\textit{e.g.}}

\def\etal{\textit{et al.~}}
\def\sArt{{state-of-the-art~}}

\newcommand{\revise}[1]{{\textcolor{black}{#1}}}

\newcommand{\figref}[1]{Fig.~\ref{#1}}
\newcommand{\tabref}[1]{Table~\ref{#1}}
\newcommand{\equref}[1]{Equ.~\ref{#1}}
\newcommand{\secref}[1]{$\S$\ref{#1}}

\ifdefined \GramaCheck
  \newcommand{\CheckRmv}[1]{}
  \renewcommand{\eqref}[1]{Equation 1}
  \renewcommand{\equref}[1]{Equation 1}
  \renewcommand{\figref}[1]{Figure 1}
  \renewcommand{\tabref}[1]{Table 1}
\else
  \newcommand{\CheckRmv}[1]{#1}
  \renewcommand{\eqref}[1]{Equation~(\ref{#1})}
\fi

\begin{document}

\title{EDN: Salient Object Detection via Extremely-Downsampled Network}

\author{Yu-Huan Wu$^*$, Yun Liu$^*$, Le Zhang,
  Ming-Ming Cheng, \textit{Senior Member, IEEE}, Bo Ren, \textit{Member, IEEE}
  \thanks{Y.-H. Wu, M.-M. Cheng and B. Ren are with TMCC, College of Computer Science, Nankai University, China. (E-mails: wuyuhuan@mail.nankai.edu.cn, cmm@nankai.edu.cn, rb@nankai.edu.cn)}
  \thanks{Y. Liu is with Computer Vision Lab, ETH Zurich, Switzerland. (E-mail: yun.liu@vision.ee.ethz.ch)}
  \thanks{L. Zhang is with University of Electronic Science and Technology of China. (E-mail: zhangleuestc@gmail.com)}
  \thanks{$^*$Y.-H. Wu and Y. Liu contributed to this work equally.
  A coin flip determines the author ordering of the first two authors.}
  \thanks{B. Ren is the corresponding author (rb@nankai.edu.cn).} \\
}

\markboth{IEEE Transactions on Image Processing, Mar. 2022}
{Wu \MakeLowercase{\textit{et al.}}: EDN: Salient Object Detection via Extremely-Downsampled Network}

\maketitle

\begin{abstract}
Recent progress on salient object detection (SOD) mainly benefits from multi-scale learning, where the high-level
and low-level features collaborate in locating salient objects and discovering fine details, respectively.
However, most efforts are devoted to low-level feature learning by fusing multi-scale features or enhancing boundary representations. High-level features, which although have long proven effective for many other tasks, yet have been barely studied for SOD.
In this paper, we tap into this gap and show that enhancing high-level features is essential for SOD as well.
To this end, we introduce an Extremely-Downsampled Network (EDN), which employs an extreme downsampling technique to effectively learn a global view of the whole image, leading to accurate salient object localization.
To accomplish better multi-level feature fusion,
we construct the Scale-Correlated Pyramid Convolution (SCPC)
to build an elegant decoder for recovering object details from the above extreme downsampling.
Extensive experiments demonstrate that EDN achieves \sArt performance with real-time speed.
Our efficient EDN-Lite also achieves competitive performance with a speed of 316fps.
Hence, this work is expected to spark some new thinking in SOD.
Code is available at \url{https://github.com/yuhuan-wu/EDN}.
\end{abstract}


\IEEEpeerreviewmaketitle

\section{Introduction}
\IEEEPARstart{S}{alient} object detection (SOD), also called saliency detection, tries to simulate the human visual system to detect the most salient and eye-catching objects or regions in natural images \cite{wang2017salient,cheng2015global, wu2021p2t}.
It has been proved to be useful for a wide range of computer vision applications such as
visual tracking \cite{mahadevan2009saliency}, scene classification \cite{ren2014region},
image retrieval \cite{gao2013visual}, and weakly supervised learning \cite{liu2020leveraging, jiang2021online}.
Much progress has been made recently
\cite{li2018contour,chen2018reverse,liu2020picanet,islam2018revisiting,he2017delving,zhao2020suppress,pang2020multi}.
However, it still remains to be challenging to detect complete salient objects in complicated scenarios accurately.

\CheckRmv{%
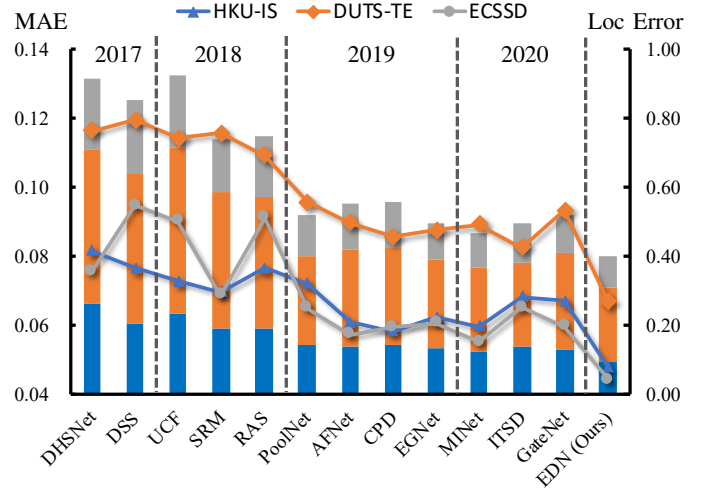
\begin{figure}[!t]
    \centering
    \begin{overpic}[width=\linewidth]{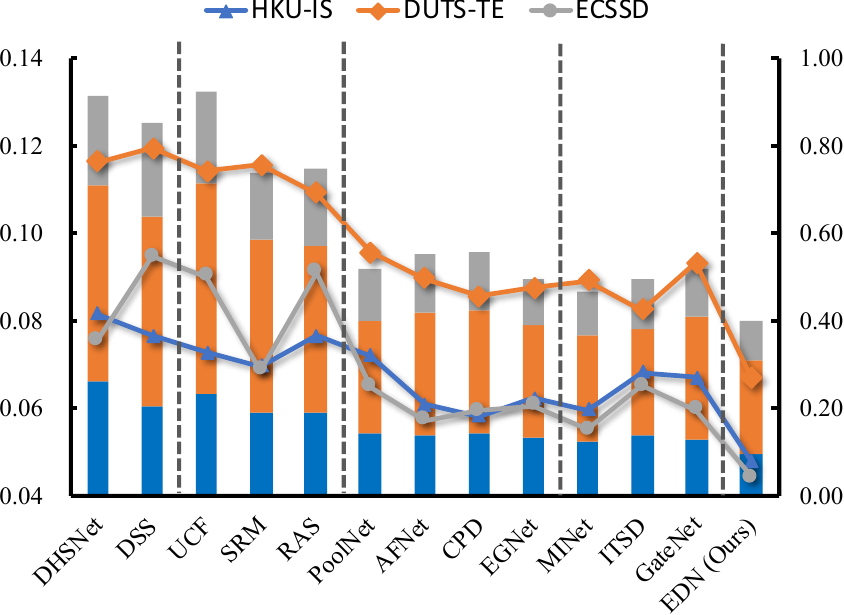}
    \put(0,69){\small{MAE}}
    \put(86,69){\small{Loc Error}}
    \put(12, 64){\small{2017}}
    \put(27, 64){\small{2018}}
    \put(50, 64){\small{2019}}
    \put(73, 64){\small{2020}}
    \end{overpic}
    \caption{\revise{MAE and cumulated localization error (Loc Error) of recent SOD methods for the center of salient objects.
    The results of MAE and Loc Error are presented by line charts and cumulated histograms, respectively.
    The above two metrics measure the capability of a method for locating salient objects.}
    Please refer to \S\ref{sec:exp_ed} for detailed experimental settings.
    From left to right, various methods are sorted by the publication date.
    It is clear to see that the accuracy for salient object localization has been saturated since 2019.}
    \label{fig:motivation}
\end{figure}%
}

In the last several years, convolutional neural networks (CNNs) have achieved vast successes in this field
\cite{luo2017non,liu2016dhsnet,wang2018salient,zhang2018bi,lee2016deep,li2017instance}.
These networks usually employ \textit{multi-scale learning} to leverage both high-level semantic features and fine-grained low-level representations, in which the former is effective in accurately locating salient objects and the latter works better in discovering object details and boundaries.
In addition, such multi-scale learning is a natural solution to
tackle the large-scale variations in practice.
Hence, most recent efforts for saliency detection are devoted to designing advanced network architectures to facilitate multi-scale learning
\cite{chen2018reverse,zeng2018learning,wang2018salient,zhang2018bi,liu2020picanet,wang2018detect,zhang2017amulet,wang2017stagewise,hou2019deeply}.

Existing multi-scale learning methods in SOD mainly aim at dealing with \textit{low-level feature learning} for better capturing/utilizing fine-grained object details/boundaries explicitly or implicitly.
For exploring fine-grained details explicitly, recent works
\cite{qin2019basnet,liu2019simple,zhao2019egnet,zhou2020interactive,wang2019salient,li2018contour,wu2019stacked,wang2017edge,feng2019attentive,su2019selectivity,wang2019focal}
try to improve the accuracy of salient object boundaries by enhancing boundary representations and imposing boundary supervision to predictions directly.
For exploring fine-grained details implicitly, many studies
\cite{zhang2018bi,wang2017stagewise,wang2018salient,liu2020picanet,islam2018revisiting,zhang2017amulet,he2017delving,li2019deep,jia2019richer,chen2018reverse}
design various multi-level feature fusion strategies to facilitate high-level semantics with low-level fine details,
for example, the hot U-Net \cite{ronneberger2015u} or the encoder-decoder based saliency detectors \cite{wang2018salient,liu2020picanet,islam2018revisiting,zhang2017amulet,liu2016dhsnet,he2017delving,li2019deep,jia2019richer}.
Many existing methods can handle object boundaries very well.
However, efforts on further performance gain have reached a bottleneck period.

To break through this bottleneck of SOD, an intuitive idea is to investigate the other aspect of multi-scale learning, \ie, \textit{high-level feature learning}, which plays an essential role in scene understanding and further locating salient objects.
\revise{Unfortunately, this direction is less investigated.
For better high-level feature learning, existing SOD methods
\cite{wang2017stagewise,zhang2018bi,liu2019simple,zhao2019pyramid,zeng2019towards}
usually directly apply some well-known modules developed
for semantic segmentation,
such as ASPP \cite{chen2017deeplab} and PSP modules \cite{zhao2017pyramid}.
However, SOD requires different high-level feature learning from semantic segmentation.
Specifically, semantic segmentation requires learning the relationship between each pixel and
all other pixels so that one can make accurate predictions according to such a relationship.
As a result, semantic segmentation methods usually aim at enlarging the receptive field to extract large-scale features for each pixel \cite{chen2017deeplab,zhao2017pyramid,huang2019ccnet,zhu2019asymmetric}.
On the other hand, SOD requires locating salient objects, which is an overall understanding of an image.
With salient object locations, object details can be easily recovered using a decoder,
like previous SOD methods that focus on low-level feature learning.}
As shown in \figref{fig:motivation}, the accuracy for locating salient objects has been saturated recently due to the limitation of high-level feature learning.
In a word, semantic segmentation needs to learn the global relationship for each pixel, while SOD needs to learn \textit{a global view of the whole image}.
Therefore, directly applying semantic segmentation methods to SOD can only achieve suboptimal performance.

To this end, this paper aims to enhance high-level feature learning, which is expected to open a new path for the future development of SOD.
We propose an \textbf{Extremely-Downsampled Block (EDB)} to learn a global view of the whole image.
EDB gradually downsamples the feature map until it becomes a feature vector, \ie, with the size of $1\times 1$.
In such a downsampling process, we keep learning deep features.
As the feature map becomes smaller, the learnt feature becomes more global.
By gradually downsampling to a feature vector, we obtain a global view of the whole image so that we can locate salient objects accurately.
The EDB only introduces a tiny computational overhead
since it operates on a very low feature resolution.
To recover complete salient objects from the global view, we build an elegant decoder to aggregate multi-level features from top to bottom gradually.
For this goal, we construct a scale-correlated pyramid convolution (SCPC) for effective feature fusion in the decoder.
Unlike traditional methods (\eg, ASPP \cite{chen2017deeplab} and PSP \cite{zhao2017pyramid}) that only adopt multiple parallel branches to extract multi-scale features \textit{separately}, SCPC adds \textit{correlation} among various branches/scales.
With EDB and effective feature fusion, the proposed \textbf{Extremely-Downsampled Network (EDN)} achieves \sArt performance on five challenging benchmarks with fast speed and a small number of parameters.
To speed up the EDN, we replace EDN's backbone with MobileNetV2 \cite{sandler2018mobilenetv2} and construct a lightweight network EDN-Lite.
It achieves competitive performance compared with recent methods with heavy backbones under the speed of 316fps.

To summarize, our contributions are as below:
\begin{itemize}
\item We propose to explore high-level feature learning for locating salient objects instead of previous low-level feature utilization for improving object boundaries, which is expected to open a new path for SOD.
\item We propose an intuitive extreme downsampling technique for learning a global view of the whole image, which generates effective high-level features for salient object localization.
%
%
\end{itemize}

\CheckRmv{%
\begin{figure*}[!t]
    \centering
    \includegraphics[width=\linewidth]{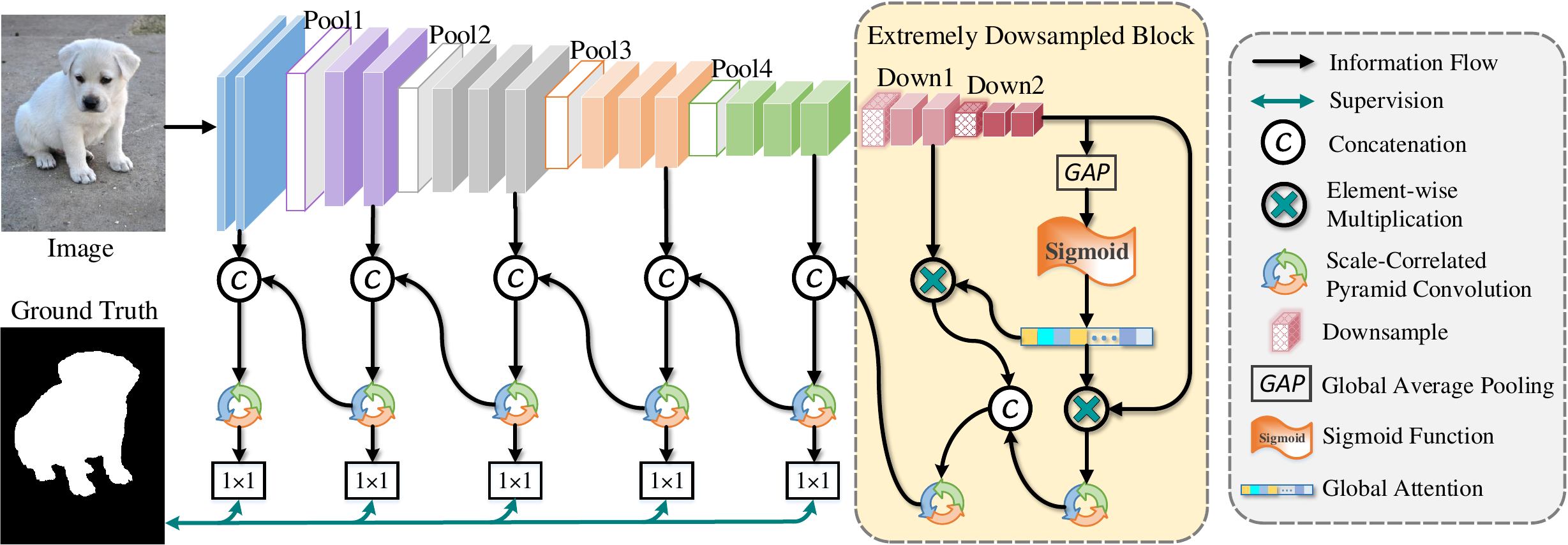} \\
    \caption{Illustration of the overall network architecture of EDN.
    We stack EDB on top of the backbone network to learn a global view of the whole image for more accurate salient object localization.
    SCPC is designed as well for effective multi-level feature integration in the decoder.
    Each downsampling operation downsamples the feature map by half except GAP.}
    \label{fig:framework}
\end{figure*}%
}

\section{Related Work} \label{sec:related}
SOD is a fundamental problem in computer vision, and thus there are a plethora of studies in the literature.

\textbf{Initiated SOD methods.}
The initiated works utilize hand-crafted features, and \revise{many shallow learning methods have been developed \cite{gong2015saliency,tu2016real,xia2017and, cong2017iterative}.}
Apart from these approaches, heuristic saliency priors also see heavy usage in this domain.
Examples include but are not limited to
color contrast \cite{cheng2015global},
center prior \cite{wang2017salient},
background prior \cite{yang2013saliency}, and so on.
However, those methods are lacking, especially compared with more recently-proposed methods,
largely due to their limited representational feature capability.


\textbf{CNN-based SOD methods.}
Inspired by the vast successes achieved by deep CNNs in other computer vision tasks, CNN-based methods have become the dominant methods for SOD.
Early CNN-based methods process and classify image regions for saliency prediction \cite{zhao2015saliency,li2015visual,lee2016deep}, which discards the spatial layout of the input image.
Motivated by the superiority of fully convolutional network (FCN)~\cite{shelhamer2017fully}, later attention has been shifted toward end-to-end image-to-image SOD
\cite{hou2019deeply,zhang2017amulet,chen2018reverse,zeng2018learning,zhang2018bi,zhao2019contrast}.
As widely acknowledged, high-level semantic features in the top CNN layers effectively locate salient objects and low-level fine-grained features in the bottom CNN layers work better in discovering object details.
Most of the recent efforts are devoted to designing effective networks to facilitate multi-scale learning.

\textbf{Multi-level feature fusion.}
Most CNN-based SOD methods achieve multi-scale learning by designing advanced network architectures for multi-level feature fusion.
The final fused features contain both high-level semantics and low-level fine details.
The architectures of these methods are usually based on
HED \cite{chen2018reverse,hou2019deeply},
Hypercolumns \cite{zeng2018learning,wang2017stagewise,su2019selectivity,zhao2019pyramid},
\revise{
or the typical U-Net \cite{wang2018salient,liu2020picanet,islam2018revisiting,zhang2017amulet,liu2016dhsnet,he2017delving,li2019deep,jia2019richer,wu2021regularized,liu2021dna,chen2021dpanet, zhang2021dense, cong2021rrnet, fang2021densely, zhang2020few, zhang2019synthesizing, zhang2020few}.}
Their target is to add low-level fine-grained features into the fused features without weakening the representation capability of high-level features, segmenting the located salient objects with clear boundaries.

\textbf{Boundary-aware SOD methods.}
Besides the above multi-level feature fusion, the recent SOD trend directly uses boundary information to improve the SOD accuracy at object boundaries
\cite{zhang2018bi,wang2017stagewise,wang2018salient,wei2020label,liu2020picanet,islam2018revisiting,zhang2017amulet,he2017delving,li2019deep,jia2019richer,chen2018reverse}.
For example, Zhao~\etal~\cite{zhao2019egnet} applied boundary supervision to low-level features.
Liu~\etal~\cite{liu2019simple} conducted joint supervision of salient objects and object boundaries at each side-output.
Zhou~\etal~\cite{zhou2020interactive} designed a two-stream network that uses two branches to learn the boundary details and locations of salient objects, respectively.

\textbf{High-level feature learning.}
While tremendous progress has been achieved, existing SOD methods mainly explore the fusion or enhancement of low-level features
to discover object boundaries better, leading to high-level feature learning less investigated.
To strengthen the high-level features, these methods
\cite{wang2017stagewise,zhang2018bi,liu2019simple,zhao2019pyramid,liu2020lightweight,zeng2019towards}
usually adopt some well-known modules developed for semantic segmentation, such as ASPP \cite{chen2017deeplab}, PSP \cite{zhao2017pyramid}, or their variants.
Due to the natural difference between SOD and semantic segmentation, as discussed above, current SOD methods can only achieve suboptimal accuracy in locating salient objects.
In this paper, we contribute from this aspect by proposing an extreme downsampling technique for better learning high-level representation in SOD.

\section{Methodology}
In this section, we first provide an overview of our method in \S\ref{sec:overveiw}.
Then, we introduce an extreme downsampling technique in \S\ref{sec:downsample}.
At last, we present the proposed SCPC and loss function in \S\ref{sec:SCPC} and \S\ref{sec:loss}, respectively.

\subsection{Overview of EDN} \label{sec:overveiw}

The overall structure of the proposed EDN is illustrated in \figref{fig:framework}.
As VGGs \cite{simonyan2014very}, ResNets \cite{he2016deep}, and MobileNets \cite{sandler2018mobilenetv2} have similar architectures of 5 stages,
without losing generality, we take VGG16 \cite{simonyan2014very} as an example backbone network to introduce EDN.
We follow previous studies \cite{liu2020picanet,wang2018salient,islam2018revisiting,li2019deep,jia2019richer, wu2021jcs} to remove the last pooling layers and all fully connected layers, resulting in an FCN \cite{shelhamer2017fully} for image-to-image saliency prediction.
So far, VGG16 has 13 convolutional layers, separated by four pooling layers.
Hence, our encoder has five convolution stages, whose outputs are denoted as $E_1$, $E_2$, $E_3$, $E_4$, and $E_5$, with scales of $1$, $\frac{1}{2}$, $\frac{1}{4}$, $\frac{1}{8}$, and $\frac{1}{16}$, respectively.

\subsubsection{High-level Feature Learning}

As discussed above, we propose an extremely-downsampled block (EDB) to learn a global view of the whole image.
By applying the EDB, we can locate salient objects accurately.
Suppose $\mathcal{F}$ denotes the function of the EDB.
We stack the EDB on top of VGG16, and the output can be written as
\CheckRmv{
\begin{equation}
    D_6 = \mathcal{F}(E_5),
\end{equation}
}
in which $D_6$ has a scale of $\frac{1}{32}$.
Here, we argue that extreme downsampling does great benefit for SOD tasks by learning a global view of the whole image.
The architecture of the EDB will be introduced in \secref{sec:downsample}.

\CheckRmv{%
\begin{figure}[!t]
    \centering
    \includegraphics[width=\linewidth]{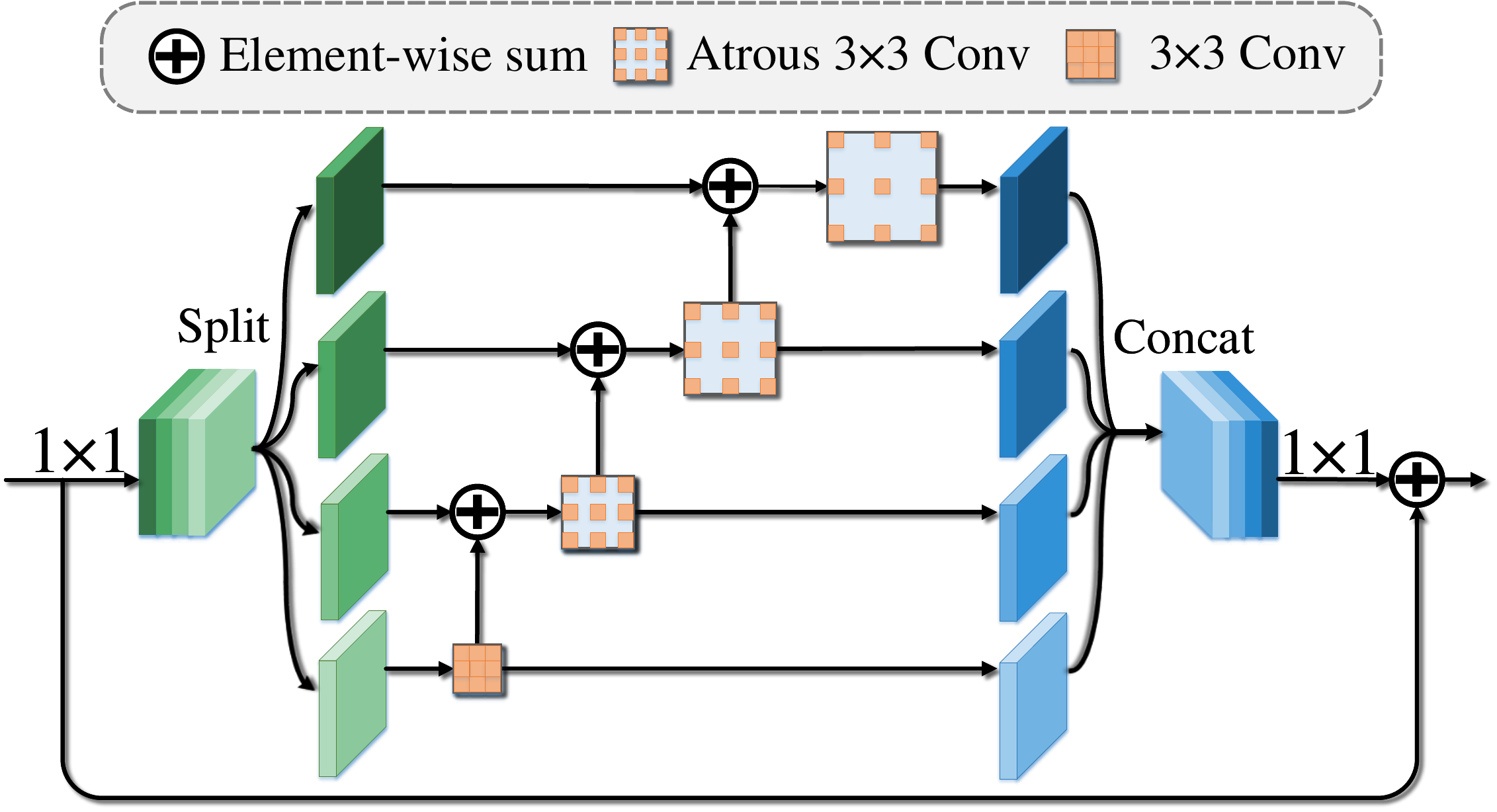}
    \caption{Illustration of SCPC for effective feature fusion.}
    \label{fig:scpc}
\end{figure}%
}

\subsubsection{Multi-level Feature Fusion}

After EDB, we perform top-down multi-level feature integration for predicting saliency maps with fine details.
To accomplish multi-level feature fusion,
we construct the scale-correlated pyramid convolution
(SCPC).
Details of the SCPC will be introduced in \secref{sec:SCPC}.
Our decoder consists of 5 fusion stages.
For each stage, we stack 2 SCPCs and let $\mathcal{H}$ denotes the function of them.
Our decoder can be elegantly formulated as
\CheckRmv{
\begin{equation}\label{equ:decoder}
\begin{aligned}
    D_{i+1}' = {\rm Upsample}({\rm Conv}_{1\times 1}(D_{i+1})),\\
    D_i = \mathcal{H}({\rm Concat}({\rm Conv}_{1\times 1}(E_i), D_{i+1}')),
\end{aligned}
\end{equation}
}
where we have $i\in\{1,2,\cdots,5\}$.
${\rm Conv}_{1\times 1}(\cdot)$ represents a $1\times 1$ convolution followed by batch normalization and ReLU layers.
${\rm Upsample}(\cdot)$ upsamples its input feature map by a scale of 2.
${\rm Concat}(\cdots)$ concatenates the input feature maps along the channel dimension.
In this way, we can effectively fuse multi-level features in an elegant way and obtain decoder outputs $D_1$, $D_2$, $D_3$, $D_4$, $D_5$, and $D_6$.

\subsection{Extremely-Downsampled Block} \label{sec:downsample}
In the above, we have discussed that existing SOD methods only focus on
learning or utilizing low-level features but ignore high-level feature learning.
Hence, we propose EDB to strengthen high-level features by learning a global view of the whole image, which leads to more accurate salient object localization (as shown in \figref{fig:motivation}).
In this part, we clarify the design details of EDB.

Suppose that the input of an EDB is $X$.
We first design a simple downsampling block to downsample the input feature map by a factor of 2 (``Down1'' in \figref{fig:framework}).
This can be formulated as
\CheckRmv{
\begin{equation}
\label{equ:downsample}
    X_1 = {\rm Conv}_{3\times 3}({\rm Conv}_{3\times 3}({\rm Downsample}(X))),
\end{equation}
}
where ${\rm Downsample}(\cdot)$ downsamples the input by a factor of 2.
${\rm Conv}_{3\times 3}(\cdot)$ is a $3\times 3$ convolution with 256 output channels, followed by batch normalization and ReLU activation.
We repeat this block to get $X_2$ (``Down2'' in \figref{fig:framework}).
$X_2$ is in a small scale and thus has a very large receptive field.
To get a global view of the whole image, we further downsample $X_2$ into a feature vector using global average pooling (GAP), which can be written as
\CheckRmv{
\begin{equation} \label{equ:global_att}
    X_3 = \sigma({\rm GAP}(X_2)).
\end{equation}
}
The value range of $X_3$ is squeezed into $[0,1]$ using a sigmoid function.
Although $X_3$ is a global representation of the input image, its size of a single pixel makes it unsuitable to start decoding from it.
Instead, we adopt it as a self-attention to recalibrate $X_2$ as
\CheckRmv{
\begin{equation}
\label{equ:calibration}
    X_2'=X_2\odot X_3,
\end{equation}
}
in which \revise{$\odot$ represents element-wise multiplication} and $X_3$ is replicated into the same size as $X_2$ before multiplication.
We also adopt $X_3$ as a nonself-attention to recalibrate $X_1$, \revise{like \equref{equ:calibration}}.
In this way, $X_1'$ and $X_2'$ are enhanced by the global representation.
Then, we fuse $X_1'$ and $X_2'$, which can be formulated as
\CheckRmv{
\begin{equation}
\begin{aligned}
    X_2'' &= {\rm Upsample}({\rm Conv}_{1\times 1}(\mathcal{H}(X_2'))),\\
    Y\ &= \mathcal{H}({\rm Concat}({\rm Conv}_{1\times 1}(X_1'), X_2'')),
\end{aligned}
\end{equation}
}
where $Y$ is the output, \ie, $Y=\mathcal{F}(X)$.
$Y$ is expected to be equipped with a global view of the whole image for better locating salient objects.

\subsection{Scale-Correlated Pyramid Convolution} \label{sec:SCPC}
We construct the SCPC for better fusing multi-level features, which is also an important aspect of multi-scale learning.
Our motivation comes from that existing modules usually
conduct \textit{separate} multi-scale feature extraction.
For example, ASPP \cite{chen2017deeplab}, PSP \cite{zhao2017pyramid},
and their numerous variants use \textit{separate} branches to extract multi-scale features,
with different branches responsible for different feature scales.
An intuitive idea is that the feature extraction at different scales should be correlated and benefit from each other.
Suppose that $M$ represents the input of SCPC.
We first apply a $1\times 1$ convolution for transition as
\CheckRmv{
\begin{equation}
    M_1 = {\rm Conv}_{1\times 1}(M).
\end{equation}
}
Then, $M_1$ is split into four feature maps evenly along the channel dimension, \ie,
\CheckRmv{
\begin{equation}
    M_2^1, M_2^2, M_2^3, M_2^4 = {\rm Split}(M_1).
\end{equation}
}
Next, we conduct multi-scale learning in a scale-correlated way, which can be formulated as
\CheckRmv{
\begin{equation}\label{equ:atrous}
\begin{aligned}
    M_3^1 &= {\rm Conv}_{3\times 3}^{a_1}(M_2^1),\\
    M_3^i &= {\rm Conv}_{3\times 3}^{a_i}(M_2^i + M_3^{i-1}),\quad i\in \{2,3,4\},
\end{aligned}
\end{equation}
}
in which ${\rm Conv}_{3\times 3}^{a_i}(\cdot)$ is a $3\times 3$ atrous convolution with an atrous rate of $a_i$.
At last, we concatenate multi-scale features and add a residual connection, like
\CheckRmv{
\begin{equation}\label{equ:SCPC}
    O = {\rm Conv}_{1\times 1}({\rm Concat}(M_3^1, M_3^2, M_3^3, M_3^4)) + M,
\end{equation}
}
where $O$ is the output, \ie, $O=\mathcal{H}(M)$.
All convolutions in SCPC are followed by batch normalization and ReLU activation, except that \equref{equ:SCPC} puts the ReLU of $1\times 1$ convolution after the residual sum with $M$, as commonly used \cite{he2016deep}.
\revise{In this way, SCPC shares the similar connections with the basic Res2Net block \cite{gao2021res2net}. The difference is that SCPC enhances the multi-scale representation learning by utilizing atrous convolutions. Specifically, SCPC learns scale-correlated features effectively by using small-scale features (with small atrous rates) to fill the holes of large-scale features (with large atrous rates) gradually through \equref{equ:atrous}.}

\subsection{Loss Function}
\label{sec:loss}

We continue by introducing our loss function for optimizing the proposed EDN.
Let $\mathcal{L}$ stands for the combination of commonly-used binary cross-entropy loss $\mathcal{L}_{bce}$ and Dice loss $\mathcal{L}_{dice}$ \cite{milletari2016v}, which can be defined as
\CheckRmv{
\begin{equation}
\begin{aligned}
    \mathcal{L}_{bce}(P,G)\ &= G\log P + (1 - G)\log (1-P),\\
    \mathcal{L}_{dice}(P,G) &= 1 - \frac{2 \cdot G\cdot P}{||G|| + ||P||},\\
    \mathcal{L}(P,G)\ \ &= \mathcal{L}_{bce} + \mathcal{L}_{dice},
\end{aligned}
\end{equation}
}
where $P$ and $G$ denote the predicted and ground-truth saliency map, respectively.
``$\cdot$'' operation indicates the dot product.
$||\cdot||$ denotes the $\ell_{1}$ norm.
The Dice loss is known as an effective way to alleviate the class imbalance of foreground and background.
The total loss for training EDN can be calculated as
\CheckRmv{
\begin{equation} \label{equ:deep_supv}
\begin{aligned}
    P_i &= \sigma({\rm Upsample}({\rm Conv}_{1\times 1}(D_i))),\\
    L\ &= \sum_{i=1}^5 \mathcal{L}(P_i, G),
\end{aligned}
\end{equation}
}
in which ${\rm Conv}_{1\times 1}(\cdot)$ does not have batch normalization and ReLU activation.
${\rm Upsample}(\cdot)$ upsamples the prediction into the size of the input image.
$\sigma(\cdot)$ is the standard sigmoid function.
We do not use $D_6$ in \equref{equ:deep_supv} due to its small size.
During testing, $P_1$ is viewed as the final output prediction of EDN.

\section{Effect of Extreme Downsampling} \label{sec:exp_ed}
Before experiments, we first discuss the effects of the proposed extreme downsampling technique.
In the above, we have clarified that existing SOD methods mainly focus on learning or better utilizing low-level fined-grained features to facilitate multi-scale learning.
However, this paper explores another direction of multi-scale learning by enhancing high-level feature learning, \ie, learning a global view of the whole image.
Here, we statistically show the benefits of extreme downsampling.
To this end, we divide the foreground of the ground-truth saliency map into boundaries, center regions, and other regions.
Boundaries are foreground pixels whose Euclidean distance to the nearest background pixel is smaller than 5 pixels, while center regions cover foreground pixels whose Euclidean distance to the nearest background pixel is in the top 20\%.
Other regions refer to foreground regions other than boundaries and center regions.
Some visualization examples of such division are displayed in the 3$^{\rm rd}$ column of \figref{fig:ed_visual_cmp}.

\newcommand{\AddImg}[1]{\includegraphics[width=.195\columnwidth]{examplesED/#1}}

\CheckRmv{%
\begin{figure}[!t]
    \centering
    \footnotesize
    \renewcommand{\arraystretch}{0.6}
    \setlength{\tabcolsep}{0.35mm}
    \begin{tabular}{ccccc}
        \AddImg{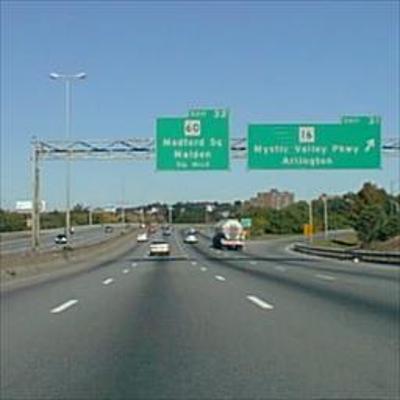} & \AddImg{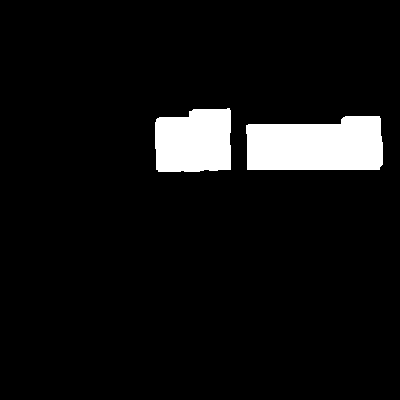}
        & \AddImg{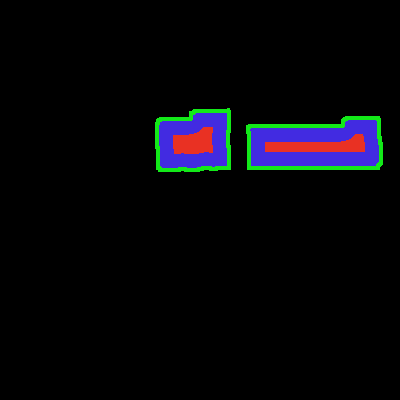} & \AddImg{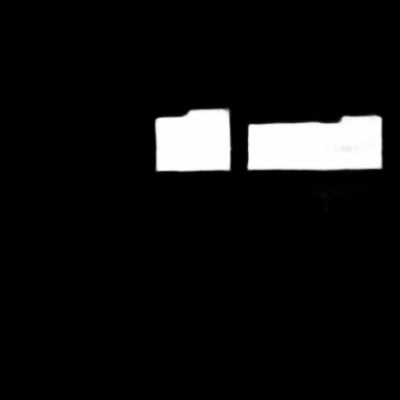}
        & \AddImg{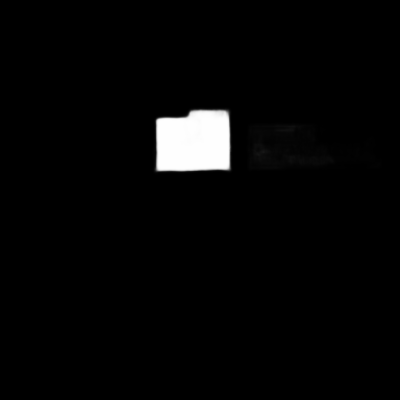}
        \\
        \AddImg{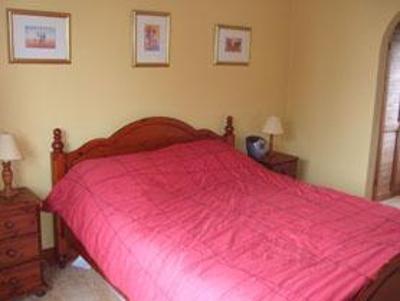} & \AddImg{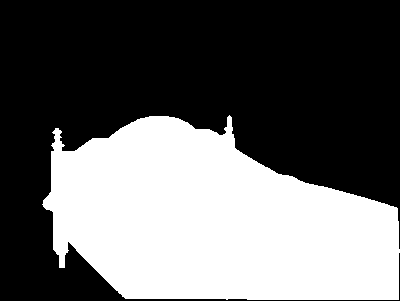}
        & \AddImg{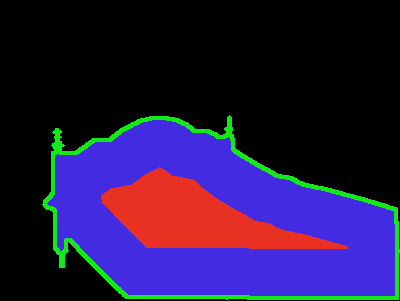} & \AddImg{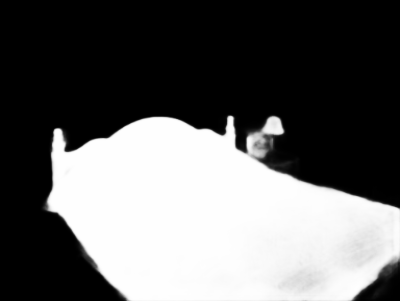}
        & \AddImg{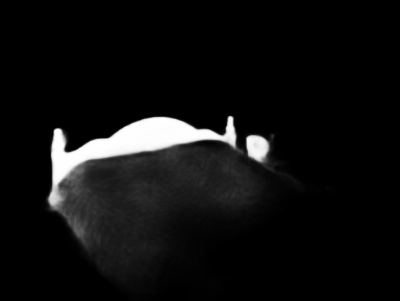}
        \\
        \AddImg{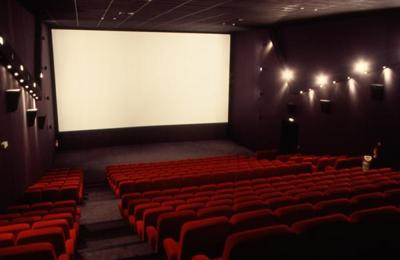} & \AddImg{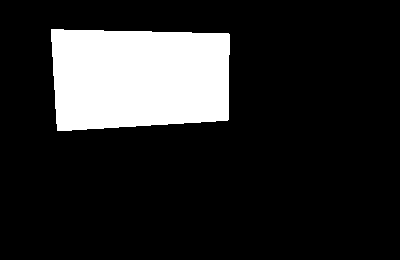}
        & \AddImg{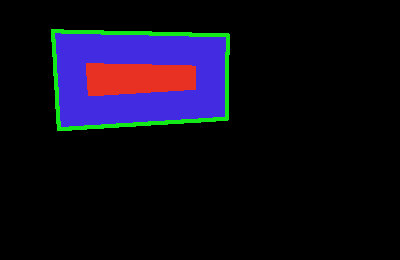} & \AddImg{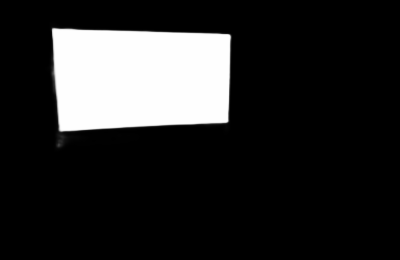}
        & \AddImg{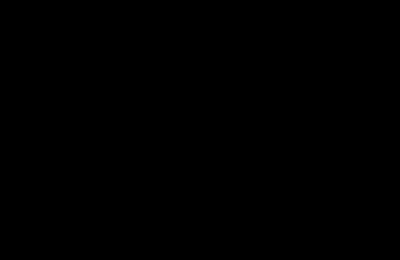}
        \\
        \AddImg{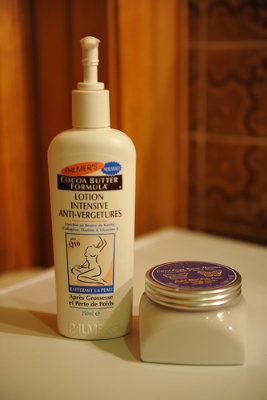} & \AddImg{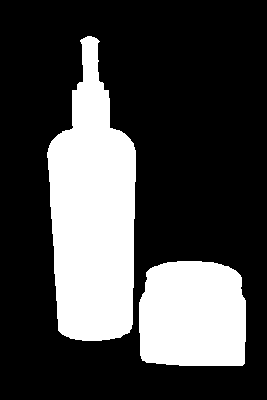} & \AddImg{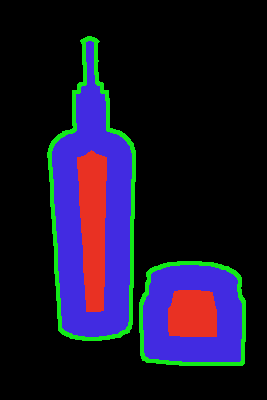}
        & \AddImg{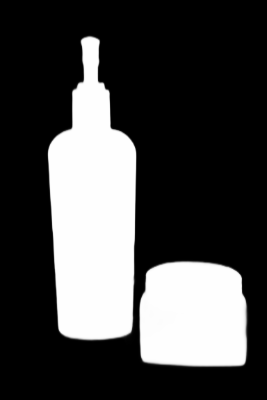} & \AddImg{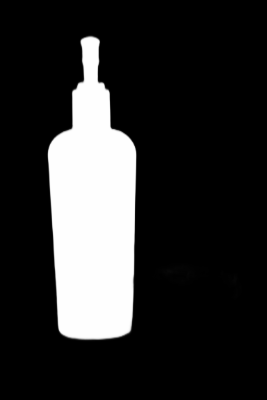}
        \\
        \AddImg{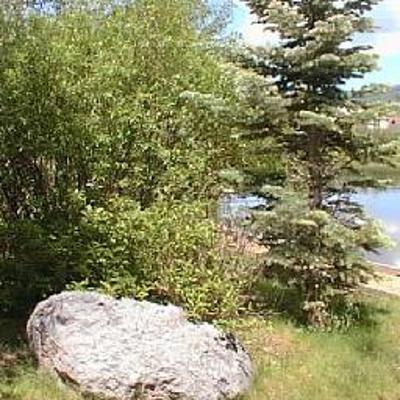} & \AddImg{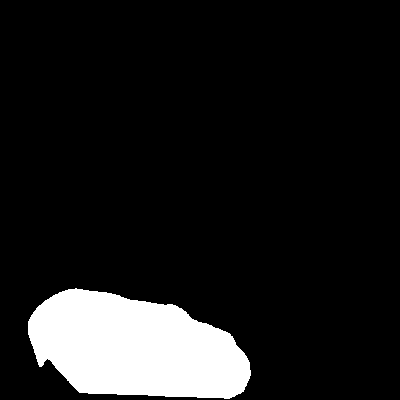}
        & \AddImg{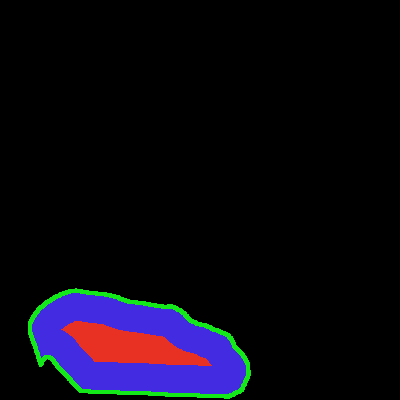} & \AddImg{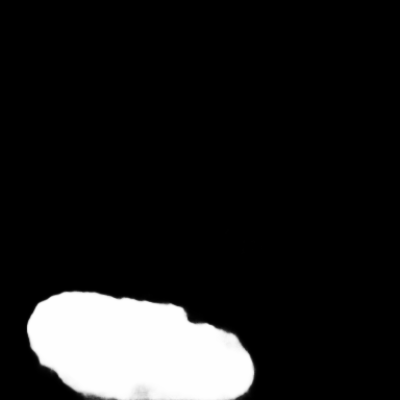}
        & \AddImg{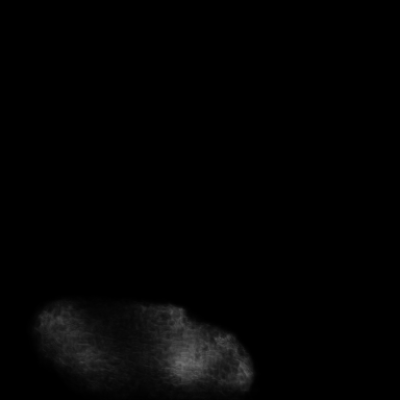}
        \\
        Image & GT & Colored GT & w/ EDB & w/o EDB
    \end{tabular}
    \caption{Visualization examples of our method with or without EDB.
    \textcolor{red}{Red}, \textcolor{green}{green}, and \textcolor{blue}{blue} pixels in the colored ground-truth saliency map indicate the center, boundary and other pixels of salient objects, respectively.
    GT: Ground-truth}
    \label{fig:ed_visual_cmp}
\end{figure}%
}

\CheckRmv{%
\begin{table}[!t]
    \centering
    \caption{Evaluation of the baseline with or without EDB in terms of the MAE metric.
    ``Rel. Impv.'' indicates the relative improvement after applying extreme downsampling.
    DUT-O: DUT-OMRON}
    \label{tab:ed_cmp}
    \renewcommand{\tabcolsep}{1.0mm}
    \resizebox{\columnwidth}{!}{%
    \begin{tabular}{c|c|ccccc} \Xhline{1pt}
        Setting & Type  & DUTS-TE & DUT-O & HKU-IS & ECSSD & PASCAL-S
        \\ \Xhline{1pt}
        Baseline & \multirow{3}[0]{*}{Center} & 0.084 & 0.178 & 0.053  & 0.053 & 0.110 \\
        $+$EDB &  & 0.062 & 0.124 & 0.043 & 0.039 & 0.082 \\
        Rel. Impv. &  & \textbf{26.6\%} & \textbf{30.1\%}
        & \textbf{19.3\%} & \textbf{26.4\%} & \textbf{25.7\%} \\
        \Xhline{1pt}
        Baseline & \multirow{3}[0]{*}{Boundary}  & 0.243 & 0.335 & 0.202 & 0.195 & 0.262 \\
        $+$EDB &  & 0.226 & 0.291 & 0.196 & 0.180 & 0.236 \\
        Rel. Impv. & & 7.0\% & 13.0\% & 3.1\%  & 7.7\% & 10.2\% \\
        \Xhline{1pt}
        Baseline & \multirow{3}[0]{*}{Other} & 0.093 & 0.181 & 0.073  & 0.071 & 0.141 \\
        $+$EDB &  & 0.076 & 0.133 & 0.065 & 0.055 & 0.112 \\
        Rel. Impv. & & 18.1\% & 26.2\% & 11.5\% & 22.5\%  & 20.7\% \\
        \Xhline{1pt}
    \end{tabular}}
\end{table}%
}

\CheckRmv{%
\begin{table*}[!t]
    \centering
    \caption{Comparison of EDN with \sArt SOD methods.
    The best performance in each column is highlighted in bold.}
    \label{tab:eval_fb_mae}%
    \renewcommand{\tabcolsep}{2mm}
    \renewcommand\arraystretch{1.1}
    \resizebox{\textwidth}{!}{%
    \begin{tabular}{l|c|c|ccc|ccc|ccc|ccc|ccc} \Xhline{1pt}
        \multirow{2}[0]{*}{Method} & Speed & \#Param
        & \multicolumn{3}{c|}{DUTS-TE  \cite{wang2017learning}}
        & \multicolumn{3}{c|}{DUT-OMRON \cite{yang2013saliency}}
        & \multicolumn{3}{c|}{HKU-IS \cite{li2015visual}}
        & \multicolumn{3}{c|}{ECSSD \cite{yan2013hierarchical}}
        & \multicolumn{3}{c}{PASCAL-S \cite{li2014secrets}}
        \\ \cline{4-18}
        & (FPS) & (M) & $F_\beta^{\text{max}}$ & $F_\beta^w$ & MAE
        & $F_\beta^{\text{max}}$ & $F_\beta^w$ & MAE & $F_\beta^{\text{max}}$ & $F_\beta^w$ & MAE
        & $F_\beta^{\text{max}}$ & $F_\beta^w$ & MAE & $F_\beta^{\text{max}}$ & $F_\beta^w$ & MAE
        \\ \Xhline{1pt}
        \multicolumn{18}{c}{VGG Backbone \cite{simonyan2014very}}
        \\ \Xhline{1pt}
        DHSNet \cite{liu2016dhsnet} & 10  & 0.059  & 0.807  & 0.705  & 0.066  &  -     &   -    &   -    & 0.889  & 0.816  & 0.053   & 94.04 & 0.906  & 0.841   & 0.820  & 0.731  & 0.092  \\
        ELD \cite{lee2016deep}  & 1     & 43.09 & 0.727  & 0.607  & 0.092  & 0.700  & 0.592  & 0.092  & 0.837  & 0.743  & 0.074 & 0.868  & 0.731  & 0.079    & 0.770  & 0.665  & 0.121  \\
        NLDF \cite{luo2017non}  & 18.5  & 35.49 & 0.806  & 0.710  & 0.065  & 0.753  & 0.634  & 0.080  & 0.902  & 0.838  & 0.048 & 0.905  & 0.839  & 0.063    & 0.822  & 0.732  & 0.098  \\
        DSS \cite{hou2019deeply}   & 7     & 62.23 & 0.813  & 0.700  & 0.065  & 0.760  & 0.643  & 0.074  & 0.900  & 0.821  & 0.050 & 0.908  & 0.835  & 0.062    & 0.829  & 0.742  & 0.095  \\
        Amulet \cite{zhang2017amulet} & 9.7   & 33.15 & 0.778  & 0.657  & 0.085  & 0.743  & 0.626  & 0.098  & 0.897  & 0.817  & 0.051 & 0.915  & 0.840  & 0.059    & 0.807  & 0.707  & 0.109  \\
        UCF \cite{zhang2017learning}  & 12    & 23.98 & 0.772  & 0.595  & 0.112  & 0.730  & 0.573  & 0.120  & 0.888  & 0.779  & 0.062 & 0.903  & 0.806  & 0.069    & 0.819  & 0.670  & 0.127  \\
        PiCANet \cite{liu2020picanet} & 5.6   & 32.85 & 0.745  & 0.054  & 0.766  & 0.691  & 0.068  & 0.916  & 0.847  & 0.042  & 0.926  & 0.865  & 0.047  & 0.837   & 0.852  & 0.767  & 0.078  \\
        C2S \cite{li2018contour}   & 16.7  & 137.03 & 0.811  & 0.717  & 0.062  & 0.759  & 0.663  & 0.072  & 0.898  & 0.835  & 0.046 & 0.911  & 0.854  & 0.053    & 0.843  & 0.765  & 0.081  \\
        RAS \cite{chen2018reverse}   & 20.4  & 20.13 & 0.831  & 0.739  & 0.059  & 0.785  & 0.695  & 0.063  & 0.914  & 0.849  & 0.045 & 0.920  & 0.860  & 0.055    & 0.828  & 0.735  & 0.100  \\
        PoolNet \cite{liu2019simple} & 43.1  & 52.51 & 0.866  & 0.783  & 0.043  & 0.791  & 0.710  & 0.057  & 0.925  & 0.864  & 0.037 & 0.939  & 0.735  & 0.045  & 0.863  & 0.782  & 0.073  \\
        AFNet \cite{feng2019attentive} & 28.4  & 35.98 & 0.857  & 0.784  & 0.046  & 0.784  & 0.717  & 0.057  & 0.921  & 0.869  & 0.036 & 0.935  & 0.782  & 0.042  & 0.861  & 0.797  & 0.070  \\
        CPD \cite{wu2019cascaded} & 68  & 29.23 & 0.864  & 0.799  & 0.043  & 0.794  & 0.715  & 0.057  & 0.924  & 0.879  & 0.033 & 0.936  & 0.895  & 0.040    & 0.861  & 0.796  & 0.072  \\
        EGNet \cite{zhao2019egnet} & 10.7  & 108.07 & 0.871  & 0.796  & 0.044  & 0.794  & 0.728  & 0.056  & 0.928  & 0.875  & 0.034 & 0.942  & 0.892  & 0.041    & 0.856  & 0.788  & 0.077  \\
        GateNet \cite{zhao2020suppress} & - & -  & 0.866  & 0.785  & 0.045  & 0.784  & 0.703  & 0.061  & 0.927  & 0.872  & 0.036   & 0.938  & 0.788  & 0.042   & 0.868  & 0.797  & 0.068  \\
        ITSD \cite{zhou2020interactive} & 53    & 17.08 & 0.875  & 0.813  & 0.042  & 0.802  & 0.734  & 0.063  & 0.926  & 0.881  & 0.035 & 0.939  & 0.797  & 0.040    & 0.869  & 0.811  & 0.068  \\
        MINet \cite{pang2020multi} & 22.3  & 47.56 & 0.870  & 0.812  & \textbf{0.040} & 0.780  & 0.719  & \textbf{0.057} & 0.929  & 0.889  & 0.032 & 0.942  & 0.811  & 0.037    & 0.864  & 0.808  & \textbf{0.065} \\
        EDN (Ours)  & 43.7  & 21.83 & \textbf{0.881} & \textbf{0.822} & 0.041  & \textbf{0.805} & \textbf{0.746} & \textbf{0.057} & \textbf{0.938} & \textbf{0.900} & \textbf{0.029} & \textbf{0.948} & \textbf{0.915} & \textbf{0.034}  & \textbf{0.875} & \textbf{0.815} & 0.066  \\
        \Xhline{1pt}
        \multicolumn{18}{c}{ResNet Backbone \cite{he2016deep}} \\
        \Xhline{1pt}
        SRM \cite{wang2017stagewise}   & 12.3  & 43.74 & 0.826  & 0.721  & 0.059  & 0.769  & 0.658  & 0.069  & 0.906  & 0.835  & 0.046 & 0.917  & 0.853  & 0.054    & 0.838  & 0.752  & 0.084  \\
        BRN \cite{wang2018detect}   & 3.6   & 126.35 & 0.827  & 0.774  & 0.050  & 0.774  & 0.709  & 0.062  & 0.910  & 0.875  & 0.036 & 0.922  & 0.891  & 0.041    & 0.849  & 0.795  & 0.072  \\
        CPD \cite{wu2019cascaded}& 32.4  & 47.85 & 0.865  & 0.794  & 0.043  & 0.797  & 0.719  & 0.056  & 0.925  & 0.875  & 0.034  & 0.939  & 0.898  & 0.037  & 0.859  & 0.794  & 0.071  \\
        BASNet \cite{qin2019basnet}& 36.2  & 87.06 & 0.859  & 0.802  & 0.048  & 0.805  & 0.751  & 0.056  & 0.928  & 0.889  & 0.032  & 0.942  & 0.904  & 0.037   & 0.854  & 0.793  & 0.076  \\
        PoolNet \cite{liu2019simple}& 40.5  & 68.26 & 0.874  & 0.806  & 0.040  & 0.792  & 0.729  & 0.055  & 0.930  & 0.881  & 0.033 & 0.943  & 0.896  & 0.039    & 0.862  & 0.793  & 0.075  \\
        EGNet \cite{zhao2019egnet}& 9.9   & 111.69 & 0.878  & 0.814  & 0.039  & 0.792  & 0.738  & 0.053  & 0.932  & 0.886  & 0.031 & 0.946  & 0.903  & 0.037    & 0.862  & 0.795  & 0.074  \\
        GCPANet \cite{chen2020global}& 51.7  & 67.06 & 0.881  & 0.820  & 0.038  & 0.796  & 0.734  & 0.057  & 0.935  & 0.889  & 0.032 & 0.946  & 0.903  & 0.036    & 0.865  & 0.808  & 0.063  \\
        GateNet \cite{zhao2020suppress} & - & -  & 0.883  & 0.808  & 0.040  & 0.806  & 0.729  & 0.055  & 0.931  & 0.880  & 0.034 & 0.945  & 0.894  & 0.041  & 0.869  & 0.797  & 0.068  \\
        ITSD \cite{zhou2020interactive}& 47.3  & 26.47 & 0.882  & 0.822  & 0.041  & 0.818  & 0.750  & 0.061  & 0.934  & 0.894  & 0.031 & 0.947  & 0.910  & 0.035  & 0.870  & 0.812  & 0.066  \\
        MINet \cite{pang2020multi}& 31.1  & 162.38 & 0.880  & 0.824  & 0.038  & 0.795  & 0.738  & 0.056  & 0.934  & 0.897  & 0.029  & 0.946  & 0.911  & 0.034  & 0.865  & 0.809  & 0.064  \\
        EDN (Ours) & 51.7  & 42.85 & \textbf{0.893} & \textbf{0.844} & \textbf{0.035} & \textbf{0.821} & \textbf{0.770} & \textbf{0.050} & \textbf{0.940} & \textbf{0.908} & \textbf{0.027} & \textbf{0.950} & \textbf{0.918} & \textbf{0.033}  & \textbf{0.879} & \textbf{0.827} & \textbf{0.062} \\
        \Xhline{1pt}

        \multicolumn{18}{c}{Lightweight Methods} \\
        \Xhline{1pt}
        CSNet \cite{gao2020highly} & 186   & 0.78 & 0.804  & 0.643  & 0.075 & 0.761  & 0.620  & 0.080    & 0.896  & 0.777  & 0.060  & 0.912  & 0.806  & 0.066  & 0.826  & 0.691  & 0.104  \\
    EDN-LiteEX (Ours) & 915   & 1.80  & 0.836  & 0.759  & 0.051  & \textbf{0.786}  & 0.716  & 0.059  & 0.911  & 0.857  & 0.040  & 0.922  & 0.869  & 0.050  & 0.836  & 0.755  & 0.084  \\
    EDN-Lite (Ours) & 316  & 1.80  & \textbf{0.856} & \textbf{0.789} & \textbf{0.045} & 0.783  & \textbf{0.721} & \textbf{0.058} & \textbf{0.924} & \textbf{0.879} & \textbf{0.034} & \textbf{0.934} & \textbf{0.890} & \textbf{0.043} & \textbf{0.852} & \textbf{0.788} & \textbf{0.073} \\
        \Xhline{1pt}
      \end{tabular}%
}
\end{table*}%
}

\vspace{5mm}

\CheckRmv{%
\begin{figure*}[!t]
    \centering
    \footnotesize
    \renewcommand{\arraystretch}{0.3}
    \setlength{\tabcolsep}{0.2mm}
    \begin{tabular}{ccc}
        \includegraphics[width=0.33\textwidth]{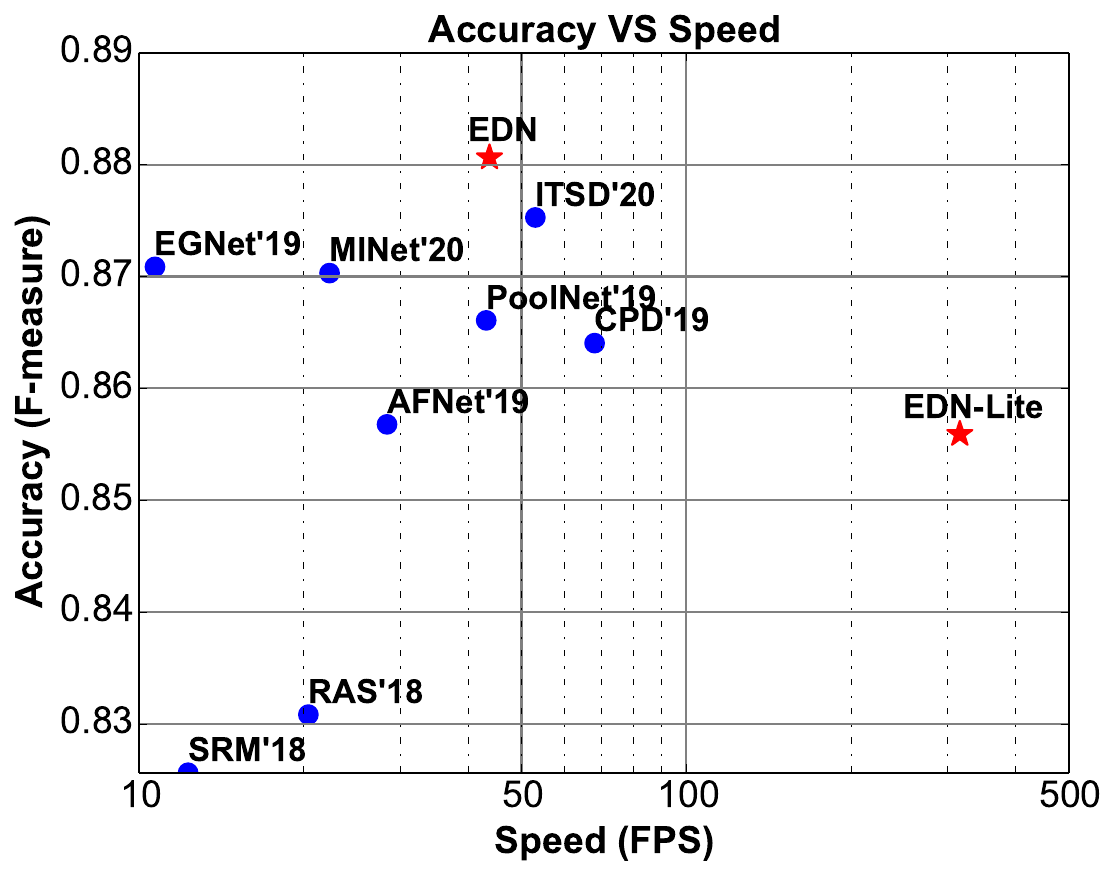} &
        \includegraphics[width=0.33\textwidth]{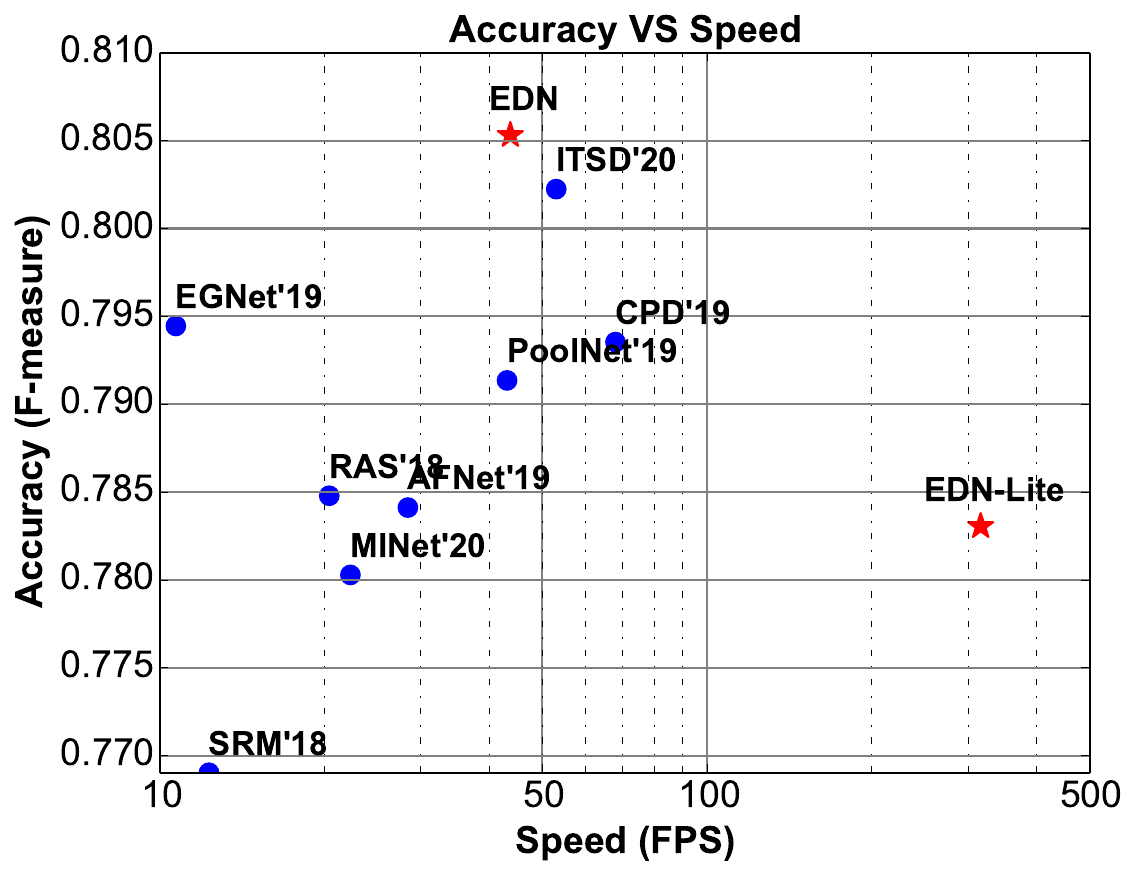} &
        \includegraphics[width=0.33\textwidth]{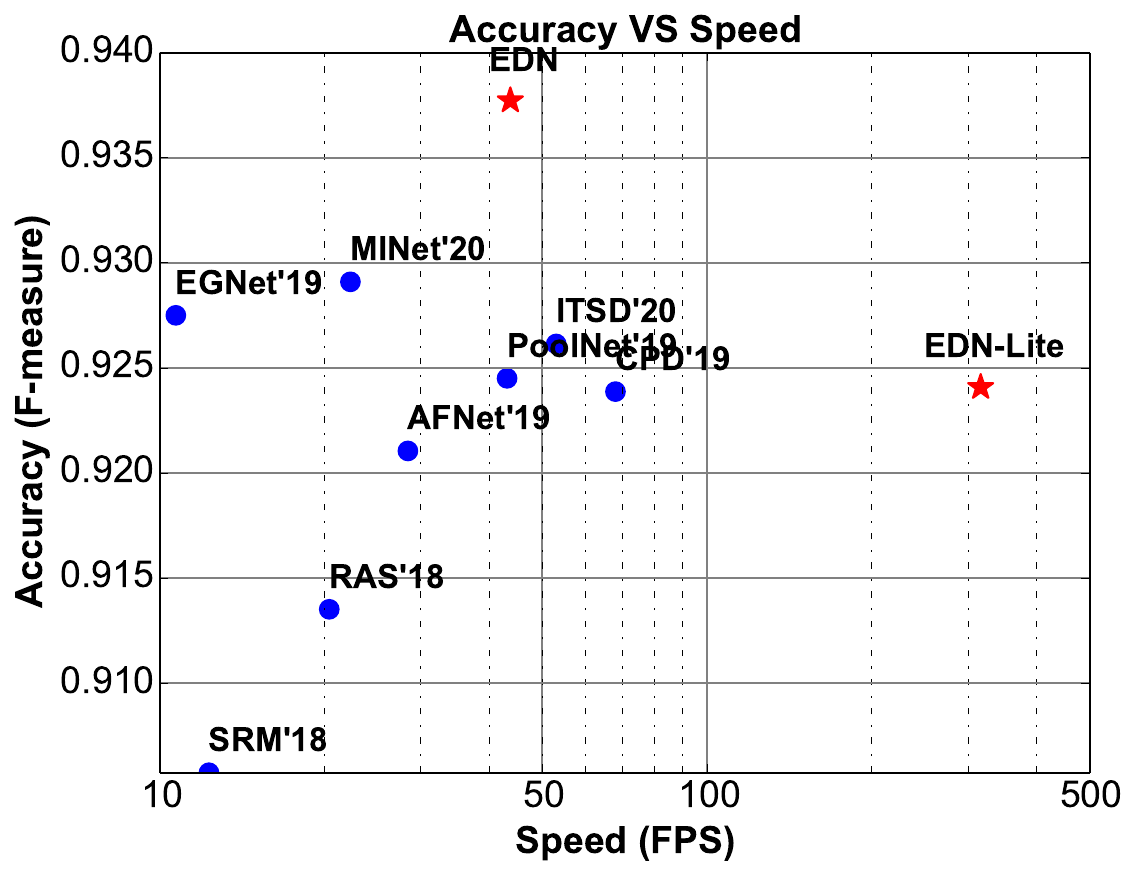}\\
        (a) DUTS-TE \cite{wang2017learning} &
        (b) DUT-OMRON \cite{liu2016dhsnet} &
        (c) HKU-IS \cite{li2015visual}
    \end{tabular}
    \caption{Speed and accuracy comparison with regular SOD methods. Our EDN largely outforms other competitors. Our EDN-Lite is also competitive compared with recent powerful SOD methods.}
    \label{fig:speed_acc}
\end{figure*}
}

\CheckRmv{%
\begin{table*}[!t]
    \centering
    \caption{Comparison of EDN with \sArt SOD methods in terms of \textbf{S-measure $S_\alpha$ \cite{fan2017structure},
    maximum E-measure $E_\xi^{\text{max}}$ \cite{fan2018enhanced}, and mean E-measure $E_\xi^{\text{mean}}$ \cite{fan2018enhanced}}.}
    \label{tab:eval_sm_em}%
    \renewcommand{\tabcolsep}{2mm}
    \renewcommand\arraystretch{1.1}
    \resizebox{\textwidth}{!}{%
    \begin{tabular}{l|c|c|ccc|ccc|ccc|ccc|ccc} \Xhline{1pt}
        \multirow{2}[0]{*}{Method} & Speed & \#Param
        & \multicolumn{3}{c|}{DUTS-TE  \cite{wang2017learning}}
        & \multicolumn{3}{c|}{DUT-OMRON \cite{yang2013saliency}}
        & \multicolumn{3}{c|}{HKU-IS \cite{li2015visual}}
        & \multicolumn{3}{c|}{ECSSD \cite{yan2013hierarchical}}
        & \multicolumn{3}{c}{PASCAL-S \cite{li2014secrets}}
        \\ \cline{4-18}
        & (FPS) & (M) & $S_\alpha$ & $E_\xi^{\text{max}}$ & $E_\xi^{\text{mean}}$
        &  $S_\alpha$ & $E_\xi^{\text{max}}$ & $E_\xi^{\text{mean}}$ &  $S_\alpha$ & $E_\xi^{\text{max}}$ & $E_\xi^{\text{mean}}$
        &  $S_\alpha$ & $E_\xi^{\text{max}}$ & $E_\xi^{\text{mean}}$ &  $S_\alpha$ & $E_\xi^{\text{max}}$ & $E_\xi^{\text{mean}}$
        \\ \Xhline{1pt}
        \multicolumn{18}{c}{VGG Backbone \cite{simonyan2014very}}
        \\ \Xhline{1pt}
        DHSNet \cite{liu2016dhsnet}& 10    & 94.04 & 0.820  & 0.880  & 0.855  & \multicolumn{1}{c}{-} & \multicolumn{1}{c}{-} & \multicolumn{1}{c}{-} & 0.870  & 0.929  & 0.905 & 0.884  & 0.928  & 0.909    & 0.810  & 0.865  & 0.845  \\
    ELD \cite{lee2016deep}   & 1     & 43.09   & 0.753  & 0.835  & 0.804  & 0.750  & 0.826  & 0.790  & 0.820  & 0.897  & 0.877 & 0.841  & 0.900  & 0.883  & 0.761  & 0.821  & 0.804  \\
    NLDF \cite{luo2017non}  & 18.5  & 35.49   & 0.816  & 0.871  & 0.852  & 0.770  & 0.820  & 0.798  & 0.879  & 0.935  & 0.914 & 0.875  & 0.922  & 0.900  & 0.805  & 0.859  & 0.844  \\
    DSS \cite{hou2019deeply}   & 7     & 62.23  & 0.826  & 0.884  & 0.851  & 0.789  & 0.842  & 0.811  & 0.881  & 0.938  & 0.907 & 0.883  & 0.927  & 0.903  & 0.809  & 0.858  & 0.847  \\
    Amulet \cite{zhang2017amulet} & 9.7   & 33.15   & 0.804  & 0.852  & 0.816  & 0.781  & 0.834  & 0.793  & 0.886  & 0.933  & 0.909 &0.894  & 0.932  & 0.909 & 0.801  & 0.847  & 0.825  \\
    UCF \cite{zhang2017learning}   & 12    & 23.98   & 0.782  & 0.844  & 0.774  & 0.760  & 0.821  & 0.760  & 0.875  & 0.926  & 0.886 & 0.884  & 0.922  & 0.890  & 0.802  & 0.855  & 0.796  \\
    PiCANet \cite{liu2020picanet}& 5.6   & 32.85  & 0.860  & 0.907  & 0.872  & 0.826  & 0.866  & 0.833  & 0.905  & 0.949  & 0.922  & 0.914  & 0.947  & 0.923   & 0.848  & 0.896  & 0.869  \\
    C2S \cite{li2018contour}   & 16.7  & 137.03  & 0.831  & 0.886  & 0.863  & 0.799  & 0.845  & 0.824  & 0.889  & 0.940  & 0.921 & 0.896  & 0.937  & 0.919   & 0.839  & 0.889  & 0.872  \\
    RAS \cite{chen2018reverse}   & 20.4  & 20.13  & 0.838  & 0.889  & 0.871  & 0.812  & 0.858  & 0.844  & 0.889  & 0.941  & 0.923 & 0.894  & 0.932  & 0.917   & 0.801  & 0.854  & 0.841  \\
    PoolNet \cite{liu2019simple}& 43.1  & 52.51  & 0.875  & 0.917  & 0.888  & 0.829  & 0.869  & 0.841  & 0.908  & 0.952  & 0.927  & 0.915  & 0.947  & 0.927  & 0.854  & 0.897  & 0.879  \\
    AFNet \cite{feng2019attentive} & 28.4  & 35.98  & 0.867  & 0.910  & 0.893  & 0.826  & 0.861  & 0.846  & 0.905  & 0.949  & 0.934 & 0.913  & 0.947  & 0.935  & 0.849  & 0.895  & 0.883  \\
    CPD \cite{wu2019cascaded}   & 68.0    & 29.23   & 0.866  & 0.911  & 0.902  & 0.818  & 0.856  & 0.845  & 0.904  & 0.948  & 0.940 & 0.910  & 0.944  & 0.938  & 0.845  & 0.888  & 0.882  \\
    EGNet \cite{zhao2019egnet}& 10.7  & 108.07    & 0.878  & 0.918  & 0.898  & 0.836  & 0.870  & 0.853  & 0.912  & 0.953  & 0.938 & 0.919  & 0.950  & 0.936  & 0.848  & 0.889  & 0.878  \\
    GateNet \cite{zhao2020suppress}& - & - & 0.870  & 0.915  & 0.893  & 0.821  & 0.858  & 0.840  & 0.910  & 0.951  & 0.934 & 0.917  & 0.948  & 0.932  & 0.857  & 0.901  & 0.886  \\
    ITSD \cite{zhou2020interactive}  & 53    & 17.08 & 0.877  & 0.919  & 0.906  & 0.829  & 0.866  & 0.853  & 0.906  & 0.950  & 0.938  &0.914  & 0.949  & 0.937   & 0.856  & 0.902  & 0.891  \\
    MINet \cite{pang2020multi}& 22.3  & 47.56  & 0.875  & 0.917  & 0.907  & 0.822  & 0.856  & 0.846  & 0.912  & 0.952  & 0.944  & 0.919  & 0.950  & 0.943   & 0.854  & 0.900  & 0.894  \\
    EDN (Ours)   & 43.7  & 21.83 & \textbf{0.883} & \textbf{0.922} & \textbf{0.912} & \textbf{0.838} & \textbf{0.871} & \textbf{0.863} & \textbf{0.921} & \textbf{0.959} & \textbf{0.950}  & \textbf{0.928} & \textbf{0.959} & \textbf{0.951} & \textbf{0.860} & \textbf{0.903} & \textbf{0.896} \\
    \Xhline{1pt}
    \multicolumn{18}{c}{ResNet Backbone \cite{he2016deep}} \\
    \Xhline{1pt}
    SRM \cite{wang2017stagewise}   & 12.3  & 43.74   & 0.836  & 0.891  & 0.854  & 0.798  & 0.844  & 0.808  & 0.887  & 0.943  & 0.913 & 0.895  & 0.937  & 0.912  & 0.834  & 0.880  & 0.857  \\
    BRN \cite{wang2018detect}   & 3.6   & 126.35   & 0.842  & 0.898  & 0.894  & 0.806  & 0.853  & 0.849  & 0.894  & 0.949  & 0.944 & 0.903  & 0.946  & 0.942 & 0.836  & 0.890  & 0.885  \\
    CPD \cite{wu2019cascaded}   & 32.4  & 47.85  & 0.869  & 0.914  & 0.898  & 0.825  & 0.868  & 0.847  & 0.905  & 0.950  & 0.938 & 0.918 & 0.951  & 0.942    & 0.848  & 0.891  & 0.882  \\
    BASNet \cite{qin2019basnet}& 36.2  & 87.06  & 0.865  & 0.903  & 0.896  & 0.836  & 0.871  & 0.865  & 0.909  & 0.951  & 0.943 & 0.916  & 0.951  & 0.943  & 0.838  & 0.886  & 0.879  \\
    PoolNet \cite{liu2019simple}& 40.5  & 68.26   & 0.883  & 0.923  & 0.904  & 0.836  & 0.871  & 0.854  & 0.915  & 0.954  & 0.939 & 0.921  & 0.952  & 0.940 & 0.849  & 0.891  & 0.880  \\
    EGNet \cite{zhao2019egnet}& 9.9   & 111.69  & 0.886  & 0.926  & 0.907  & 0.841  & 0.878  & 0.857  & 0.917  & 0.956  & 0.942 & 0.925  & 0.955  & 0.943  & 0.852  & 0.892  & 0.881  \\
    GCPANet \cite{chen2020global}& 51.7  & 67.06   & 0.890  & 0.929  & 0.912  & 0.839  & 0.868  & 0.853  & 0.920  & 0.958  & 0.945 & 0.927  & 0.955  & 0.944 & 0.864  & 0.907  & 0.895  \\
    GateNet \cite{zhao2020suppress} & - & -   & 0.885  & 0.928  & 0.906  & 0.838  & 0.876  & 0.856  & 0.915  & 0.955  & 0.938 & 0.920  & 0.952  & 0.936 & 0.858  & 0.904  & 0.887  \\
    ITSD \cite{zhou2020interactive}  & 47.3  & 26.47   & 0.884  & 0.930  & 0.914  & 0.840  & 0.880  & 0.865  & 0.917  & 0.960  & 0.947 &0.925  & \textbf{0.959} & 0.947 & 0.859  & \textbf{0.908}  & 0.895  \\
    MINet \cite{pang2020multi}& 31.1  & 162.38  & 0.883  & 0.927  & 0.917  & 0.833  & 0.869  & 0.860  & 0.919  & 0.960  & 0.952 & 0.925  & 0.957  & 0.950  & 0.856  & 0.903  & 0.896  \\
    EDN (Ours)   & 51.7    & 42.85  & \textbf{0.892} & \textbf{0.934} & \textbf{0.925} & \textbf{0.849} & \textbf{0.885} & \textbf{0.878} & \textbf{0.924} & \textbf{0.962} & \textbf{0.955} & \textbf{0.927} & 0.958  & \textbf{0.951} & \textbf{0.865} & \textbf{0.908} & \textbf{0.902} \\
    \Xhline{1pt}
    \multicolumn{18}{c}{Lightweight Methods} \\
    \Xhline{1pt}
    CSNet \cite{gao2020highly} & 186  & 0.78  & 0.822  & 0.875  & 0.820  & 0.805  & 0.853  & 0.801  & 0.881  & 0.933  & 0.883  & 0.893  & 0.931  & 0.886  & 0.814  & 0.860  & 0.815  \\
    EDN-LiteEX (Ours) & 915   & 1.80   & 0.848  & 0.903  & 0.882  & 0.823  & \textbf{0.867}  & 0.851  & 0.894  & 0.945  & 0.928  & 0.899  & 0.938  & 0.925  & 0.820  & 0.869  & 0.853  \\
    EDN-Lite (Ours) & 316  & 1.80   & \textbf{0.862} & \textbf{0.910} & \textbf{0.895} & \textbf{0.824}  & 0.861  & \textbf{0.848} & \textbf{0.907} & \textbf{0.950} & \textbf{0.938} & \textbf{0.911} & \textbf{0.944} & \textbf{0.933} & \textbf{0.842} & \textbf{0.890} & \textbf{0.878} \\
    \Xhline{1pt}

    \end{tabular}}
\end{table*}%
}

\renewcommand{\AddImg}[1]{%
    \includegraphics[width=.088\textwidth]{examples/#1.jpg} &%
    \includegraphics[width=.088\textwidth]{examples/#1.png} &%
    \includegraphics[width=.088\textwidth]{examples/#1_a} &%
    \includegraphics[width=.088\textwidth]{examples/#1_minet} &%
    \includegraphics[width=.088\textwidth]{examples/#1_itsd} &%
    \includegraphics[width=.088\textwidth]{examples/#1_egnet} &%
    \includegraphics[width=.088\textwidth]{examples/#1_cpd} &%
    \includegraphics[width=.088\textwidth]{examples/#1_ras} &%
    \includegraphics[width=.088\textwidth]{examples/#1_picanet} &%
    \includegraphics[width=.088\textwidth]{examples/#1_dss} &%
    \includegraphics[width=.088\textwidth]{examples/#1_amulet} &%
}

\CheckRmv{%
\begin{figure*}[!t]
    \centering
    \footnotesize
    \renewcommand{\arraystretch}{0.6}
    \setlength{\tabcolsep}{0.35mm}
    \begin{tabular}{ccccccccccccc}
        \AddImg{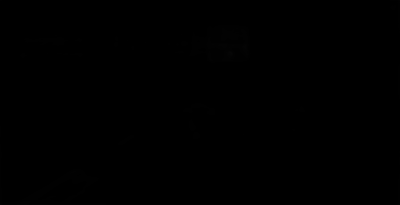} \\
        \AddImg{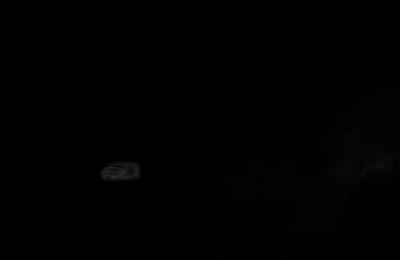} \\
        \AddImg{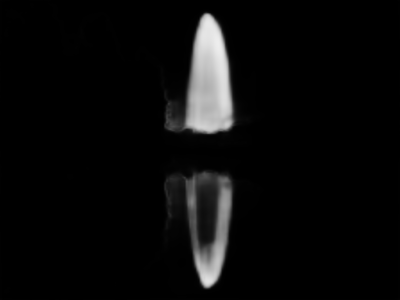} \\
        \AddImg{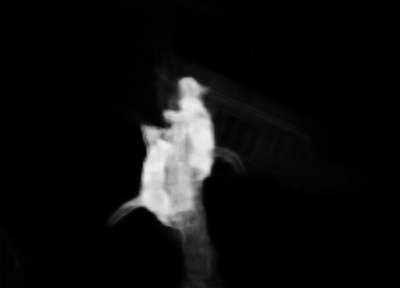} \\
        \AddImg{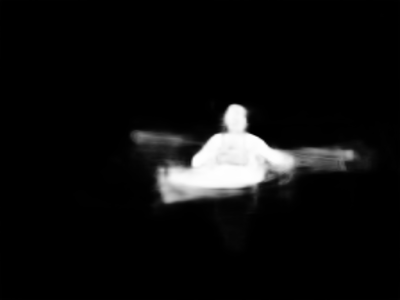} \\
        \AddImg{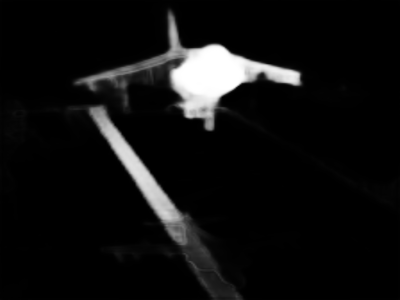} \\
        Image & GT & EDN (Ours) & MINet\cite{pang2020multi} & ITSD\cite{zhou2020interactive}
        & EGNet\cite{zhao2019egnet} & CPD\cite{wu2019cascaded} & RAS\cite{chen2018reverse}
        & PiCANet \cite{liu2020picanet} & DSS\cite{hou2019deeply} & Amulet\cite{zhang2017amulet}
    \end{tabular}
    \caption{Qualitative comparison. Our predictions are more similar to the ground-truth (GT).}
    \label{fig:examples}
\end{figure*}%
}

With the above definition, we compute the mean absolute error (MAE) for the center, boundary, and other regions, respectively.
Please see \S\ref{sec:exp_setup} for more details about the metric and datasets.
Note that when we compute MAE for one type of region,
the other two types of regions are ignored.
The statistical results are shown in \tabref{tab:ed_cmp}.
We remove EDB from the proposed EDN to serve as the baseline.
The relative improvement in \tabref{tab:ed_cmp} is the fraction of $\Delta$MAE and MAE of the baseline, where $\Delta\text{MAE}$ is the decrease of MAE by adding EDB to the baseline.
Applying EDB, we observe that the relative improvement in terms of center regions is much larger than that in terms of boundaries and other regions, which suggests that the improvement brought by EDB mainly comes from the accurate localization of salient objects.
\figref{fig:motivation} shows that salient object localization accuracy has been saturated since 2019,
while EDB boosts such accuracy significantly.
Therefore, EDB has achieved its goal of improving SOD through better salient object localization.
Moreover, it is interesting to find that EDB also has some improvement in terms of boundaries, although it is designed for high-level feature learning.
One potential reason is that powerful high-level features make the decoding process easier, leading to better utilization of low-level features.
Some visualization examples are provided in \figref{fig:ed_visual_cmp}.
EDB can help the system detect all salient objects.
Without EDB, some salient objects will be lost completely (the 1$^{\rm st}$, 3$^{\rm rd}$, and 4$^{th}$ rows) or partially (the 2$^{\rm nd}$ and 5$^{\rm th}$ rows).

\revise{Moreover, we follow \cite{russakovsky2015imagenet} to define a localization metric
that measures the accuracy of locating salient objects.
We compute the intersection-over-union (IoU) between the ground-truth and the predicted result.
If the IoU is not better than a specific threshold (0.7 as a strict threshold \cite{everingham2010pascal}), we define that the predicted result does not locate salient objects well.
Therefore, methods only with clear boundaries may not get good results on this metric 
if they are defective on the accurate localization of salient objects.
We present the comparison results between our EDN with other \sArt methods in \figref{fig:motivation}. 
As can be observed, the accuracy for salient object localization has been saturated since 2019.
}

\section{Experiments}
\subsection{Experimental Setup} \label{sec:exp_setup}
\paragraph{Implementation details}
The proposed method is implemented using the PyTorch \cite{paszke2019pytorch} and Jittor \cite{hu2020jittor} library.
The training of all experiments is conducted using the Adam \cite{kingma2015adam} optimizer with parameters $\beta_1=0.9$, $\beta_2=0.99$, weight decay $10^{-4}$, and batch size $24$.
We adopt the \textit{poly} learning rate scheduler so that the learning rate for the $n^{\rm th}$ epoch is $init\_lr \times \left( 1 - \frac{n}{max\_epoch} \right)^{power}$, where we have $init\_lr = 5 \times 10^{-5}$ and $power = 0.9$.
The training lasts for $30$ epochs in total.
In the ResNet-based EDN and MobileNetV2-based \textbf{EDN-Lite}, we replace the Conv$_{3\times 3}$
block in extreme downsampling with the bottleneck \cite{he2016deep} and inverted residual block \cite{sandler2018mobilenetv2}, respectively.
In EDN-Lite, we replace Conv$_{3\times 3}$ operations of all SCPCs with
depth-wise separable $3\times 3$ convolutions.
In training, the backbone networks of EDN and EDN-Lite are both pretrained on ImageNet, and we freeze the batch normalization layers of backbones as commonly done.
In testing, the input images are resized into $384\times 384$ for both EDN and EDN-Lite.

\paragraph{Datasets}
We extensively evaluate the proposed EDN on five datasets,
including DUTS \cite{wang2017learning},
ECSSD \cite{yan2013hierarchical},
HKU-IS \cite{li2015visual},
PASCAL-S \cite{li2014secrets},
and DUT-OMRON \cite{yang2013saliency} datasets.
These five datasets consist of $15572$, $1000$, $4447$, $850$ and $5168$ natural images with corresponding pixel-level labels.
Following recent studies \cite{wang2018detect,wang2017stagewise,liu2020picanet,zeng2018learning}, we train EDN on the DUTS training set and evaluate on the DUTS test set (DUTS-TE) and other four datasets.

\CheckRmv{%
\begin{table*}[!t]
    \centering
    \caption{Evaluation of various design choices of EDB. \revise{``GA'' and ``ED'' denote global attention and extremely downsampling, respectively.}}
    \label{tab:interior_ed}
    \renewcommand{\tabcolsep}{1.4mm}
    \resizebox{\textwidth}{!}{%
    \begin{tabular}{c|l|ccc|ccc|ccc|ccc|ccc} \Xhline{1pt}
        \multirow{2}[0]{*}{No.} & \multirow{2}[0]{*}{Method}
        & \multicolumn{3}{c|}{DUTS-TE \cite{wang2017learning}}
        & \multicolumn{3}{c|}{DUT-OMRON \cite{yang2013saliency}}
        & \multicolumn{3}{c|}{HKU-IS \cite{li2015visual}}
        & \multicolumn{3}{c|}{ECSSD \cite{yan2013hierarchical}}
        & \multicolumn{3}{c}{PASCAL-S \cite{li2014secrets}}
        \\ \cline{3-17}
        & & $F_\beta^{\text{max}}$ & $F_\beta^w$ & MAE & $F_\beta^{\text{max}}$ & $F_\beta^w$ & MAE
        & $F_\beta^{\text{max}}$ & $F_\beta^w$ & MAE & $F_\beta^{\text{max}}$ & $F_\beta^w$ & MAE
        & $F_\beta^{\text{max}}$ & $F_\beta^w$ & MAE
        \\ \Xhline{1pt}
        1 & Backbone & 0.779 & 0.691 & 0.065
        & 0.682 & 0.573 & 0.094 & 0.883 & 0.819 & 0.049 & 0.886  & 0.816  & 0.068  & 0.814 & 0.733 & 0.091
        \\
        2 & No. 1$+$Decoder & 0.871 & 0.816 & \textbf{0.039}
        & 0.780 & 0.725 & \textbf{0.054} & 0.932 & 0.896 & 0.030 & 0.938  & 0.904  & 0.037   & 0.864 & 0.806 & 0.068
        \\
        3 & No. 2$+$\revise{EDB (w/ 1 block)} & 0.874 & 0.820 & 0.041
        & 0.794 & 0.741 & 0.056 & 0.934 & 0.899 & \textbf{0.029} & 0.943  & 0.911  & 0.035  & 0.871 & 0.818 & \textbf{0.064}
        \\
        4 & No. 2$+$\revise{EDB (w/o GA)} & 0.876 & \textbf{0.822} & 0.041 & 0.803 & \textbf{0.747} & 0.056
        & 0.936 & \textbf{0.901} & \textbf{0.029} & 0.944  & 0.912  & 0.035  & 0.873 & \textbf{0.819} & 0.066
        \\
        5 & No. 2$+$\revise{EDB (w/o ED)} & 0.861 & 0.797 & 0.047 & 0.798 & 0.728 & 0.062
        & 0.931 & 0.893 & 0.031 & 0.941  & 0.905  & 0.036  & 0.865 & 0.805 & 0.068
        \\
        6 & No. 2$+$\revise{EDB (default)} & \textbf{0.881} & \textbf{0.822} & 0.041 & \textbf{0.805} & 0.746 & 0.057
        & \textbf{0.938} & 0.900 & \textbf{0.029} & \textbf{0.948} & \textbf{0.915} & \textbf{0.034}
         & \textbf{0.875} & 0.815 & 0.066
        \\ \Xhline{1pt}
    \end{tabular}}
\end{table*}%
}

\CheckRmv{%
\begin{table*}[!t]
    \centering
    \caption{Comparison of EDB with other alternatives on five datasets.}
    \label{tab:ed_design}
    \renewcommand{\tabcolsep}{2mm}
    \resizebox{\textwidth}{!}{%
    \begin{tabular}{l|ccc|ccc|ccc|ccc|ccc} \Xhline{1pt}
         \multirow{2}[0]{*}{Method}
        & \multicolumn{3}{c|}{DUTS-TE \cite{wang2017learning}}
        & \multicolumn{3}{c|}{DUT-OMRON \cite{yang2013saliency}}
        & \multicolumn{3}{c|}{HKU-IS \cite{li2015visual}}
        & \multicolumn{3}{c|}{ECSSD \cite{yan2013hierarchical}}
        & \multicolumn{3}{c}{PASCAL-S \cite{li2014secrets}}
        \\ \cline{2-16}
         & $F_\beta^{\text{max}}$ & $F_\beta^w$ & MAE & $F_\beta^{\text{max}}$ & $F_\beta^w$ & MAE
        & $F_\beta^{\text{max}}$ & $F_\beta^w$ & MAE & $F_\beta^{\text{max}}$ & $F_\beta^w$ & MAE
        & $F_\beta^{\text{max}}$ & $F_\beta^w$ & MAE
        \\ \Xhline{1pt}
         Baseline   & 0.871  & 0.816  & \textbf{0.039}  & 0.780  & 0.725  & 0.054  & 0.932  & 0.896  & 0.030 & 0.938  & 0.904  & 0.037  & 0.864  & 0.806  & 0.068  \\
     $+$ASPP~\cite{chen2017deeplab}   & 0.873  & 0.816  & \textbf{0.039}  & 0.790  & 0.735  & \textbf{0.053}  & 0.933  & 0.893  & 0.031  & 0.940  & 0.902  & 0.039  & 0.856  & 0.800  & 0.070  \\
    $+$PSP~\cite{zhao2017pyramid}    & 0.870  & 0.812  & 0.042  & 0.789  & 0.732  & 0.056  & 0.934  & 0.898  & 0.030 & 0.939  & 0.901  & 0.038  & 0.869  & 0.810  & 0.068  \\
     $+$NL~\cite{wang2018non}   & 0.869  & 0.815  & 0.040  & 0.784  & 0.725  & 0.055  & 0.931  & 0.896  & 0.030 & 0.936  & 0.902  & 0.037  & 0.870  & 0.809  & 0.068  \\
     $+$DenseASPP~\cite{yang2018denseaspp}   & 0.866  & 0.813  & 0.040  & 0.775  & 0.721  & 0.056  & 0.930  & 0.895  & 0.029 & 0.936  & 0.899  & 0.038 & 0.864  & 0.808  & 0.065  \\
     $+$EDB  & \textbf{0.881} & \textbf{0.822} & 0.041  & \textbf{0.805} & \textbf{0.746} & 0.057  & \textbf{0.938} & \textbf{0.900} & \textbf{0.029}  & \textbf{0.948} & \textbf{0.915} & \textbf{0.034} & \textbf{0.875} & \textbf{0.815} & \textbf{0.066} \\
\Xhline{1pt}
    \end{tabular}}
\end{table*}%
}

\begin{table*}[!t]
    \centering
    \caption{\revise{Comparison of the default channel-wise global attention with other alternatives on five datasets.}}
    \label{tab:global_att_abla}%
    \resizebox{\textwidth}{!}{%
      \begin{tabular}{l|ccc|ccc|ccc|ccc|ccc}\Xhline{1pt}
      \multicolumn{1}{c|}{\multirow{2}[0]{*}{Method}}
      & \multicolumn{3}{c|}{DUTS-TE \cite{wang2017learning}} & \multicolumn{3}{c|}{DUT-OMRON \cite{yang2013saliency}} & \multicolumn{3}{c|}{HKU-IS \cite{li2015visual}}
      & \multicolumn{3}{c|}{ECSSD \cite{yan2013hierarchical}}
      & \multicolumn{3}{c}{PASCAL-S \cite{li2014secrets}}
      \\ \cline{2-16}
            & $F_\beta^{\text{max}}$ & $F_\beta^w$ & MAE & $F_\beta^{\text{max}}$ & $F_\beta^w$ & MAE & $F_\beta^{\text{max}}$ & $F_\beta^w$ & MAE & $F_\beta^{\text{max}}$ & $F_\beta^w$ & MAE & $F_\beta^{\text{max}}$ & $F_\beta^w$ & MAE \\
            \Xhline{1pt}
      Matrix Multiply    & 0.874 & 0.819 & 0.042 & 0.798 & 0.740 & 0.059 &
      0.936 & \textbf{0.902} & \textbf{0.029} & 0.944 & 0.913 & 0.035 & 0.872 & 0.815 & 0.066  \\
      Spatial-wise   & 0.876 & 0.821 & \textbf{0.041} & 0.802 & 0.742 & \textbf{0.056} & 0.937 & \textbf{0.902} & \textbf{0.029} & 0.942 & 0.909 & 0.036
      & 0.873 & \textbf{0.817} & \textbf{0.065} \\
      Default  & \textbf{0.881} & \textbf{0.822} & \textbf{0.041}  & \textbf{0.805} & \textbf{0.746} & 0.057  & \textbf{0.938} & 0.900 & \textbf{0.029}  & \textbf{0.948} & \textbf{0.915} & \textbf{0.034} & \textbf{0.875} & 0.815 & 0.066 \\
      \Xhline{1pt}
    \end{tabular}}
\end{table*}%

\paragraph{Evaluation criteria}
We evaluate EDN against previous state-of-the-art methods with regard to three widely-used metrics, \ie, ${\rm F}$-measure score ($F_{\beta}$), mean absolute error ({\rm MAE}), and weighted ${\rm F}$-measure score ($F_\beta^w$).
For the first metric, ${\rm F}$-measure is the weighted harmonic mean of precision and recall, like
\begin{equation}
F_{\beta} = \frac{(1 + \beta^2) \times {\rm Precision} \times {\rm Recall}}{\beta^2 \times {\rm Precision} + {\rm Recall}},
\end{equation}
where we set $\beta^2 = 0.3$ to emphasize the importance of precision, following previous works \cite{hou2019deeply,liu2019simple,liu2020picanet,zhang2017amulet}.
In our paper, we report the maximum $F_{\beta}$, \ie, $F_\beta^{\text{max}}$, under different binarizing thresholds.
Higher ${\rm F}$-measure indicates better performance.
The second metric, MAE, measures the similarity between the predicted saliency map $P$ and the ground-truth saliency map $G$, which can be computed as
\begin{equation}
{\rm MAE}(P, G) = \frac{1}{HW} \sum_{i=1}^{H} \sum_{j=1}^{W} \left| P_{i,j} - G_{i,j} \right|,
\end{equation}
where $H$ and $W$ denote the height and width of the saliency map, respectively.
The lower the MAE is, the better the SOD method is.
The third metric, weighted ${\rm F}$-measure $F_\beta^w$, solves the problems of ${\rm F}$-measure that may cause interpolation flaw, dependency flaw, and equal-importance flaw \cite{margolin2014evaluate}.
We use the official code with the default setting of the authors to conduct the evaluation.
The higher the weighted ${\rm F}$-measure is, the better the performance is.

Recently, S-measure \cite{fan2017structure} and E-measure \cite{fan2018enhanced} have been widely applied for SOD evaluation in many works \cite{pang2020multi,zhao2020depth}.
Following them, we also compare our method with others using these two metrics.
S-measure calculates the structural similarity between the predicted saliency map
and the ground-truth map.
E-measure computes the similarity for the binarized predicted map and the binary ground-truth map.
Here, we compute the maximum and average E-measure among all thresholds that binarize the predicted map.
We use the official code to compute the scores of S-measure and E-measure.
More details about these two measures can refer to the corresponding original papers \cite{fan2017structure,fan2018enhanced}.

\subsection{Comparison with State-of-the-art Methods} \label{sec:sart_cmp}
In this part, we compare the proposed EDN with existing 20 recent regular methods, including
DHSNet \cite{liu2016dhsnet}, ELD \cite{lee2016deep}, NLDF \cite{luo2017non},
DSS \cite{hou2019deeply}, Amulet \cite{zhang2017amulet}, UCF \cite{zhang2017learning},
PiCANet \cite{liu2020picanet}, C2S \cite{li2018contour}, RAS \cite{chen2018reverse},
PoolNet \cite{liu2019simple}, AFNet \cite{feng2019attentive}, CPD \cite{wu2019cascaded},
EGNet \cite{zhao2019egnet}, GateNet \cite{zhao2020suppress}, ITSD \cite{zhou2020interactive},
MINet \cite{zhou2020interactive}, BRN \cite{wang2018detect}, SRM \cite{wang2017stagewise},
BASNet \cite{qin2019basnet}, and GCPANet \cite{chen2020global}.
We evaluate them using both VGG16~\cite{simonyan2014very} and ResNet-50~\cite{he2016deep} backbones.
We also compare our MobileNetV2-based EDN-Lite with the very recent lightweight SOD method CSNet \cite{gao2020highly}.
To further speed up EDN-Lite, we construct EDN-LiteEX which is the EDN-Lite tested with a smaller input size ($224 \times 224$).
Since DHSNet \cite{liu2016dhsnet} uses DUT-OMRON \cite{yang2013saliency} for training, we do not report its result on the DUT-OMRON \cite{yang2013saliency} dataset.
For a fair comparison, we use the saliency maps provided by the original authors and if not provided, we directly use their official code and models to compute the missing saliency maps.
We also report each method's speed and number of parameters for reference.
The speed is tested using each method's official code and a single NVIDIA TITAN Xp GPU.

\paragraph{Quantitative comparison}
We show the results in \tabref{tab:eval_fb_mae} (F-measure, weighted F-measure, and MAE)
and \tabref{tab:eval_sm_em} (S-measure, maximum E-measure, and mean E-measure).
We also visualize the speed and accuracy comparison on the three largest datasets, \ie, DUTS-TE, DUT-OMRON, and HKU-IS, in \figref{fig:speed_acc}.
EDN consistently achieves the best performance in most cases,
and in a few remaining cases, EDN is also very close to the best performance.
EDN also has real-time speed and a relatively small number of parameters.
EDN's lightweight version, \ie, EDN-Lite, achieves competitive performance compared with recent \sArt methods with $10\times$ speed on average,
while the other lightweight method CSNet \cite{gao2020highly} still has a large performance gap compared with recent \sArt methods.
The above results demonstrate the efficacy and efficiency of EDN and EDN-Lite.

\paragraph{Qualitative comparison}
The qualitative comparison is displayed in \figref{fig:examples}.
While other competitors may not detect the whole salient objects
or even not find some salient objects in difficult scenarios,
EDN can segment salient objects with clear boundaries.

\CheckRmv{%
\begin{table*}[!t]
    \centering
    \caption{Evaluation of various atrous rate settings for SCPC. The last row is the default setting.}
    \label{tab:atrous_rate}
    \renewcommand{\tabcolsep}{2.5mm}
    \resizebox{1.0\textwidth}{!}{%
    \begin{tabular}{c|ccc|ccc|ccc|ccc|ccc|ccc} \Xhline{1pt}
        \multirow{2}[0]{*}{No.} & \multicolumn{3}{c|}{Setting}
        & \multicolumn{3}{c|}{DUTS-TE \cite{wang2017learning}}
        & \multicolumn{3}{c|}{DUT-OMRON \cite{yang2013saliency}}
        & \multicolumn{3}{c|}{HKU-IS \cite{li2015visual}}
        & \multicolumn{3}{c|}{ECSSD \cite{yan2013hierarchical}}
        & \multicolumn{3}{c}{PASCAL-S \cite{li2014secrets}}
        \\ \cline{2-19}
        & L & H & EH & $F_\beta^{\text{max}}$ & $F_\beta^w$ & MAE & $F_\beta^{\text{max}}$ & $F_\beta^w$ & MAE & $F_\beta^{\text{max}}$
        & $F_\beta^w$ & MAE & $F_\beta^{\text{max}}$ & $F_\beta^w$ & MAE & $F_\beta^{\text{max}}$ & $F_\beta^w$ & MAE
        \\ \Xhline{1pt}
        1 & (b) & - & -  & 0.877 & 0.824 & 0.041 & 0.802 & 0.745 & 0.057
        & 0.937 & \textbf{0.901} & \textbf{0.029} & 0.945  & 0.911  & 0.035 & 0.870 & 0.813 & 0.067
        \\
        2 & (c) & - & -& 0.880 & \textbf{0.825} & 0.040 & 0.804 & \textbf{0.750}
        & \textbf{0.054} & 0.935  & 0.899 & 0.030  & 0.945  & 0.913  & \textbf{0.034}  & 0.874 & \textbf{0.822} & \textbf{0.064}
        \\
        3 & - & (a) & -  & 0.875 & 0.820 & 0.042 & 0.798
        & 0.740 & 0.059 & 0.935 & 0.900 & \textbf{0.029} & 0.945  & 0.913  & 0.035 & 0.869 & 0.814 & 0.067
        \\
        4 & - & (c) & -  & 0.878 & 0.824 & \textbf{0.039} & 0.800 & 0.747
        & \textbf{0.054} & 0.935 & 0.900  & \textbf{0.029} & 0.946  & 0.913  & 0.036 & 0.873 & 0.818 & 0.066
        \\
        5 & - & - & (a) & 0.873 & 0.809 & 0.044 & 0.803 & 0.741 & 0.059
        & 0.933 & 0.893 & 0.032 & 0.943  & 0.906  & 0.038  & 0.872 & 0.813 & 0.068
        \\
        6 & - & - & (b)  & 0.873 & 0.811 & 0.045 & 0.801 & 0.737 & 0.061
        & 0.935 & 0.896 & 0.030 & 0.947  & 0.910  & 0.036 & 0.870 & 0.811 & 0.070
        \\
        7 & - & - & -  & \textbf{0.881} & 0.822 & 0.041
        & \textbf{0.805} & 0.746 & 0.057  & \textbf{0.938} & 0.900 & \textbf{0.029} & \textbf{0.948} & \textbf{0.915} & \textbf{0.034}
        & \textbf{0.875} & 0.815 & 0.066\\
        \Xhline{1pt}
    \end{tabular}}
\end{table*}}

\begin{table*}[!t]
    \centering
    \caption{Comparison of SCPC with other alternatives on five datasets.}
    \label{tab:scpc_cmp}%
    \resizebox{\textwidth}{!}{%
      \begin{tabular}{l|ccc|ccc|ccc|ccc|ccc}\Xhline{1pt}
      \multicolumn{1}{c|}{\multirow{2}[0]{*}{Method}}
      & \multicolumn{3}{c|}{DUTS-TE \cite{wang2017learning}} & \multicolumn{3}{c|}{DUT-OMRON \cite{yang2013saliency}} & \multicolumn{3}{c|}{HKU-IS \cite{li2015visual}}
      & \multicolumn{3}{c|}{ECSSD \cite{yan2013hierarchical}}
      & \multicolumn{3}{c}{PASCAL-S \cite{li2014secrets}}
      \\ \cline{2-16}
            & $F_\beta^{\text{max}}$ & $F_\beta^w$ & MAE & $F_\beta^{\text{max}}$ & $F_\beta^w$ & MAE & $F_\beta^{\text{max}}$ & $F_\beta^w$ & MAE & $F_\beta^{\text{max}}$ & $F_\beta^w$ & MAE & $F_\beta^{\text{max}}$ & $F_\beta^w$ & MAE \\
            \Xhline{1pt}
      Conv    & 0.837  & 0.776  & 0.048  & 0.740  & 0.662  & 0.070  & 0.919  & 0.880  & 0.034 & 0.924  & 0.876  & 0.049  & 0.855  & 0.790  & 0.074  \\
      ASPP~\cite{chen2017deeplab}   & 0.864  & 0.805  & 0.042  & 0.774  & 0.712  & 0.057  & 0.929  & 0.890  & 0.032   & 0.935  & 0.899  & 0.040 & \textbf{0.868} & \textbf{0.806} & \textbf{0.068} \\
      SCPC   & \textbf{0.871} & \textbf{0.816} & \textbf{\textbf{0.039}} & \textbf{0.780} & \textbf{0.725} & \textbf{0.054} & \textbf{0.932} & \textbf{0.896} & \textbf{0.030} & \textbf{0.938} & \textbf{0.904} & \textbf{0.037} & 0.864  & \textbf{0.806} & \textbf{0.068} \\
      \Xhline{1pt}
      \end{tabular}}
  \end{table*}%

\begin{table*}[htbp]
    \centering
    \caption{Performance with different loss functions.}
    \resizebox{1.0\textwidth}{!}{%
      \begin{tabular}{c|c|ccc|ccc|ccc|ccc|ccc}
        \Xhline{1pt}
        \multicolumn{1}{c|}{\multirow{2}[0]{*}{No.}}    & \multicolumn{1}{c|}{\multirow{2}[0]{*}{Loss Setting}}  & \multicolumn{3}{c|}{DUTS-TE \cite{wang2017learning}}
         & \multicolumn{3}{c|}{DUT-OMRON \cite{liu2016dhsnet}}
         & \multicolumn{3}{c|}{HKU-IS \cite{li2015visual}}
         & \multicolumn{3}{c|}{ECSSD \cite{yan2013hierarchical}}
         & \multicolumn{3}{c}{PASCAL-S \cite{li2014secrets}} \\
        \cline{3-17}
       &  & $F_\beta^{\text{max}}$ & $F_\beta^w$ & MAE & $F_\beta^{\text{max}}$ & $F_\beta^w$ & MAE
        & $F_\beta^{\text{max}}$ & $F_\beta^w$ & MAE & $F_\beta^{\text{max}}$ & $F_\beta^w$ & MAE
        & $F_\beta^{\text{max}}$ & $F_\beta^w$ & MAE\\
      \Xhline{1pt}
      1     & BCE only & 0.880  & 0.817  & 0.041  & 0.803  & 0.741  & 0.059  & 0.936  & 0.895  & 0.030  & 0.946  & 0.908  & 0.037  & \textbf{0.875} & 0.817  & 0.066  \\
      2     & Dice only & 0.876  & \textbf{0.835} & \textbf{0.038} & 0.802  & \textbf{0.757} & \textbf{0.054}  & 0.934  & \textbf{0.907} & \textbf{0.028} & 0.946  & \textbf{0.921} & \textbf{0.033} & 0.868  & \textbf{0.819} & \textbf{0.065} \\
      3     & BCE + Dice & \textbf{0.881} & 0.822  & 0.041  & \textbf{0.805} & 0.746  & 0.057  & \textbf{0.938} & 0.900  & 0.029  & \textbf{0.948} & 0.915  & 0.034  & \textbf{0.875} & 0.815  & 0.066  \\
      \Xhline{1pt}
      \end{tabular}%
  }
    \label{tab:loss}%
  \end{table*}%

\subsection{Ablation Study} \label{ablation}
In this section, we conduct ablation study for EDN equipped with the proposed EDB and SCPC.
All experiments in this part are based on the VGG-16 backbone \cite{simonyan2014very}.
Other settings are the same as \S\ref{sec:exp_setup}.

\paragraph{Effect of various design choices for EDB}
Other than showing the effect of the whole EDB in \S\ref{sec:exp_ed},
we conduct analyses on the interior design choices of EDB.
More specifically, we control the number of downsampling operations and the allowance of global attention to the output features in EDB.
The results are summarized in \tabref{tab:interior_ed}.
``Backbone'' means to predict saliency maps directly from the last stage of the VGG16 backbone.
\revise{``EDB (w/ 1 block)'' indicates EDB with only one downsampling block (Down1 in \figref{fig:framework}).
``EDB (w/o GA)'' indicates EDB without global attention (\equref{equ:global_att} - \equref{equ:calibration}).
``EDB (w/o ED)'' only removes the downsampling operations but remains all convolutions and global attention.
As can be seen, ``EDB (default)'' outperforms ``EDB w/o GA'', showing that global attention is significant in EDB.
Besides, ``EDB (default)'' substantially outperforms ``EDB (w/o ED)'' and the baseline without EDB.
This demonstrates the significance of the downsampling and global attention in EDB, and removing each element will affect the performance significantly.}

\paragraph{Comparison of EDB with other alternatives}
Here, we replace EDB with other modules for high-level feature learning, like ASPP \cite{chen2017deeplab}, PSP \cite{zhao2017pyramid}, Non-local (NL) \cite{wang2018non}, and DenseASPP \cite{yang2018denseaspp} modules.
ASPP,  PSP, and DenseASPP modules perform multi-scale feature learning using multiple separate branches.
The results are shown in \tabref{tab:ed_design}.
We can find that adding ASPP, PSP, NL, or DenseASPP module to the baseline only achieves slightly better or even worse performance.
In contrast, EDB outperforms ASPP, PSP, NL, DenseASPP, and the baseline by a large margin, demonstrating the superiority of our extreme downsampling technique.

\paragraph{Choices of global attention}
\revise{As described in \secref{sec:downsample},
we apply channel-wise element-wise multiplication as the default strategy for
global attention. To validate the effectiveness of this choice,
we perform ablation study using spatial attention or matrix multiplication instead.
The results are shown in \tabref{tab:global_att_abla}.
We can observe that both spatial attention and matrix multiplication have worse
performance than the default strategy.
Therefore, the default channel-wise element-wise multiplication is the best choice.
}

\paragraph{Atrous rate configurations of SCPC}
EDN has six downsampling operations, downsampling the feature map by half each time.
Correspondingly, there are seven SCPC modules whose atrous rates are set according to the size of the feature map, as shown in \figref{fig:framework}.
We show the results of different atrous rate settings for SCPC in \tabref{tab:atrous_rate}.
We divide our seven times of multi-level feature fusion into 3 groups.
``L'' (low) includes the first two stages that output feature maps with the highest resolutions.
``H'' (high) includes the 3$^{\rm rd}$, 4$^{\rm th}$, and 5$^{\rm th}$ stages.
``EH'' (extremely high) includes the last two extra scales of feature maps in EDB.
For different groups, we apply different atrous rate settings.
By default, the atrous rates of four branches in SCPC for the group ``L'', ``H'', and ``EH'' are set as \{1, 2, 4, 8\} (a), \{1, 2, 3, 4\} (b), and \{1, 1, 1, 1\} (c), respectively.
In \tabref{tab:atrous_rate}, we tried two other types of atrous rate settings for each group.
We can observe that the results only fluctuate slightly with various atrous rates, demonstrating that the proposed SCPC is robust for different atrous rate settings.
Since the 7$^{\rm th}$ setting in \tabref{tab:atrous_rate} achieves the overall best performance, we employ it as the default setting for SCPC.

\paragraph{Comparing SCPC with other alternatives}
In this part, we compare the proposed SCPC with the vanilla convolution (``Conv'') and ASPP.
Specifically, we first replace SCPC with $3\times 3$ convolutions that have the same number of output channels as SCPC, resulting in a decoder similar to U-Net \cite{ronneberger2015u}.
Then, we replace SCPC with ASPP by removing the scale correlation in SCPC, \ie, removing the sum term of $M_3^{i-1}$ in \equref{equ:atrous}.
The results are displayed in \tabref{tab:scpc_cmp}.
We can see that ASPP outperforms ``Conv'' significantly, and SCPC further improves ASPP substantially, suggesting the effectiveness of SCPC in feature fusion.

\paragraph{Discussion of the loss function}
As a default, we use the hybrid loss which consists of the binary
cross-entropy (BCE) loss and Dice loss.
To validate this design choice, we also test the performance of training
with  a single loss function (BCE loss only or Dice loss only).
Results are shown in \tabref{tab:loss}.
As can be observed, the Dice loss can help improve $F_\beta^w$ and MAE
but decreases the score of $F_{\beta}$.
Since $F_{\beta}$ is known as the primary metric in SOD, we apply the hybrid of the BCE loss and Dice loss as our default setting.

\renewcommand{\AddImg}[1]{%
    \includegraphics[width=.11\textwidth]{failcases/2365#1} &%
    \includegraphics[width=.11\textwidth]{failcases/5729#1} &%
    \includegraphics[width=.11\textwidth]{failcases/sun_aaqopqrpmqlxfvyh#1} &%
    \includegraphics[width=.11\textwidth]{failcases/sun_akhpeprzgvznapdl#1} %
}

\CheckRmv{%
\begin{figure}[!t]
    \centering
    \footnotesize
    \renewcommand{\arraystretch}{0.2}
    \setlength{\tabcolsep}{0.35mm}
    \begin{tabular}{cccccc}
        \rotatebox[origin=l]{90}{\hspace{4.5mm} Image} & \AddImg{.jpg} \\
        \rotatebox[origin=l]{90}{\hspace{5.3mm} GT}& \AddImg{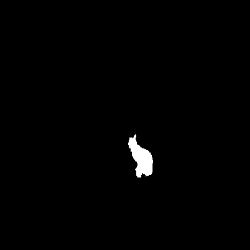} \\
        \rotatebox[origin=l]{90}{\hspace{5.1mm} Ours} & \AddImg{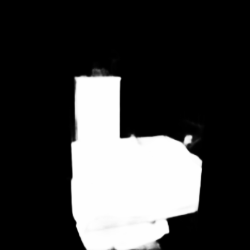} \\
        & No. 1 & No. 2 & No. 3 & No. 4 \\
    \end{tabular}
    \caption{Representative failure cases of the proposed EDN.}
    \label{fig:failure_cases}
\end{figure}%
}

\subsection{\revise{Discussions about Failure Cases}}

\revise{Although the proposed EDN achieves great success in global view learning for SOD, there is still large room for further improvement.
We present some representative failure cases in \figref{fig:failure_cases}.
As can be seen, EDN fails in some confusing scenarios. 
For example, EDN may predict wrong salient regions (No. 1 in \figref{fig:failure_cases}).
EDN may predict the largest salient object but not the most discriminative salient object (No. 2, 3 in \figref{fig:failure_cases}).
EDN may regard discriminative lanes as non-salient regions (No. 4 in \figref{fig:failure_cases}).
Even so, the improvement in both quantitative and qualitative comparisons in \secref{sec:sart_cmp} demonstrates that EDN can deal with most scenarios well and achieve the new \sArt for SOD.
}

\section{Conclusion}
In SOD, high-level semantic features are effective for salient object localization and low-level fine details captures the object boundaries well~\cite{chen2018reverse,zeng2018learning,wang2018salient,zhang2018bi,liu2020picanet,wang2018detect,zhang2017amulet,wang2017stagewise,hou2019deeply, wu2021mobilesal, wu2021regularized}. This observation has sparked extensive studies on enhancing low-level features
\cite{zhang2018bi,wang2017stagewise,wang2018salient,liu2020picanet,islam2018revisiting,zhang2017amulet,he2017delving,li2019deep,jia2019richer,chen2018reverse,liu2016dhsnet,qin2019basnet,liu2019simple,zhao2019egnet,zhou2020interactive,wang2019salient,li2018contour,wu2019stacked,wang2017edge,feng2019attentive,su2019selectivity,wang2019focal} but interestingly, high-level feature learning is barely investigated. We tap into the gap by proposing an extremely-downsampled block (EDB) to learn a better global view of the whole image and thus accurately localize salient object. A scale-correlated pyramid convolution (SCPC) strategy is also proposed to build an elegant and effective decoder to recover object details from the above extreme downsampling.
This work could be served as a strong baseline for SOD and spark new efforts towards enhancing high-level features as well.

{\small
\bibliographystyle{IEEEtran}
\bibliography{reference}
}

\newcommand{\AddPhoto}[1]{\vspace{-.1in} \includegraphics[width=1in]{#1}}

\begin{IEEEbiography}[\vspace{-.1in} \AddPhoto{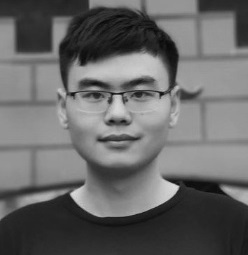}]{Yu-Huan Wu}
    received the bachelor's degree from Xidian University, in 2018. He is currently pursuing the Ph.D. degree with the College of Computer Science, Nankai University, supervised by Prof. M.-M. Cheng. His research interests include computer vision and machine learning.
\end{IEEEbiography}

\begin{IEEEbiography}[\vspace{-.1in} \AddPhoto{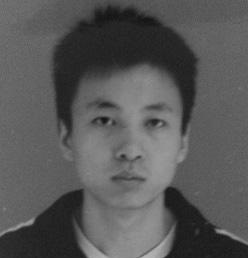}]{Yun Liu}
received his bachelor's and doctoral degrees from Nankai University
in 2016 and 2020, respectively.
Currently, he works with Prof. Luc Van Gool as a postdoctoral scholar at ETH Zurich.
His research interests include computer vision and machine learning.
\end{IEEEbiography}

\begin{IEEEbiography}[\AddPhoto{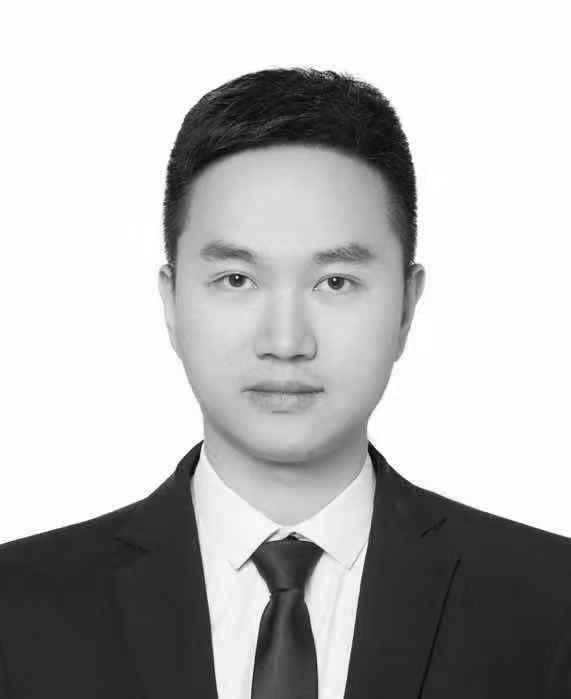}]{Le Zhang}
received his M.Sc and Ph.D.degree form Nanyang Technological University
(NTU) in 2012 and 2016, respectively.
Currently, he is a scientist at Institute for Infocomm Research,
Agency for Science, Technology and Research (A*STAR), Singapore.
He served as TPC member in several conferences such as AAAI, IJCAI.
He has served as a Guest
Editor for Pattern Recognition and Neurocomputing;
His current research interests include deep learning and computer vision.
\end{IEEEbiography}

\begin{IEEEbiography}[\AddPhoto{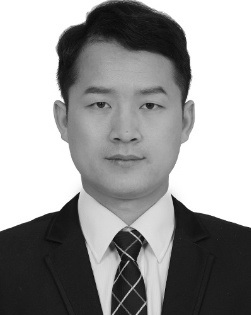}]{Ming-Ming Cheng}
received his PhD degree from Tsinghua University in 2012.
Then he did two years research fellow with Prof. Philip Torr
in Oxford.
He is now a professor at Nankai University, leading the
Media Computing Lab.
His research interests include computer graphics, computer
vision, and image processing.
He received research awards, including ACM China Rising Star Award,
IBM Global SUR Award, and CCF-Intel Young Faculty Researcher Program.
He is on the editorial boards of IEEE TIP.
\end{IEEEbiography}

\begin{IEEEbiography}[\AddPhoto{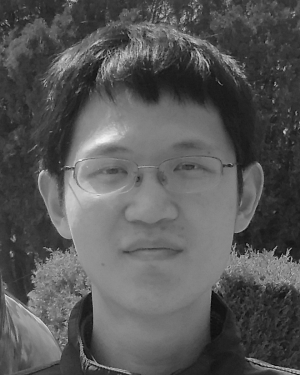}]{Bo Ren}
received the PhD degree from Tsinghua University in 2015. He
is currently an associate professor in the College of Computer Science, Nankai University, Tianjin. His research interests include computer graphics, computer vision, and image processing.
\end{IEEEbiography}

\vfill

\end{document}